\documentclass[letterpaper]{article} % DO NOT CHANGE THIS
\usepackage{aaai23}  % DO NOT CHANGE THIS
\usepackage{times}  % DO NOT CHANGE THIS
\usepackage{helvet}  % DO NOT CHANGE THIS
\usepackage{courier}  % DO NOT CHANGE THIS
\usepackage[hyphens]{url}  % DO NOT CHANGE THIS
\usepackage{graphicx} % DO NOT CHANGE THIS
\usepackage{natbib}  % DO NOT CHANGE THIS AND DO NOT ADD ANY OPTIONS TO IT
\usepackage{caption} % DO NOT CHANGE THIS AND DO NOT ADD ANY OPTIONS TO IT
\usepackage{amsfonts}
\usepackage{algorithm}
\usepackage{algorithmicx}
\usepackage{algpseudocode}
\usepackage{amsmath}

\usepackage{subcaption}
\usepackage{babel}
\usepackage{tabu}
\usepackage{wrapfig,lipsum,booktabs}
\usepackage{stfloats}
\usepackage{kantlipsum}
\usepackage{booktabs}
\usepackage{newfloat}
\usepackage{listings}

\DeclareCaptionStyle{ruled}{labelfont=normalfont,labelsep=colon,strut=off} % DO NOT CHANGE THIS
\floatstyle{ruled}
\newfloat{listing}{tb}{lst}{}
\floatname{listing}{Listing}

\newcommand{\thickhline}{%
    \noalign {\ifnum 0=`}\fi \hrule height 1pt
    \futurelet \reserved@a \@xhline
}
\newcolumntype{"}{@{\hskip\tabcolsep\vrule width 1pt\hskip\tabcolsep}}

\newcommand{\PreserveBackslash}[1]{\let\temp=\\#1\let\\=\temp}
\newcolumntype{C}[1]{>{\PreserveBackslash\centering}p{#1}}
\newcolumntype{R}[1]{>{\PreserveBackslash\raggedleft}p{#1}}
\newcolumntype{L}[1]{>{\PreserveBackslash\raggedright}p{#1}}

\newtheorem{manualtheoreminner}{Theorem}
\newenvironment{manualtheorem}[1]{%
  \renewcommand\themanualtheoreminner{#1}%
  \manualtheoreminner
}{\endmanualtheoreminner}

\title{Open-Ended Diverse Solution Discovery with Regulated Behavior Patterns for Cross-Domain Adaptation}
\author{
    Kang Xu,
    Yan Ma,
    Bingsheng Wei,
    Wei Li \thanks{Corresponding Author}
}
\affiliations{
    Academy for Engineering and Technology, Fudan University, Shanghai, China\\
    kangxu21@m.fudan.edu.cn,~
    \{20210860024,~weibingsheng,~fd\_liwei\}@fudan.edu.cn
}

\usepackage{bibentry}
\begin{document}

\maketitle

\begin{abstract}
While Reinforcement Learning can achieve impressive results for complex tasks, the learned policies are generally prone to fail in downstream tasks with even minor model mismatch or unexpected perturbations. Recent works have demonstrated that a policy population with diverse behavior characteristics can generalize to downstream environments with various discrepancies. However, such policies might result in catastrophic damage during the deployment in practical scenarios like real-world systems due to the unrestricted behaviors of trained policies. Furthermore, training diverse policies without regulation of the behavior can result in inadequate feasible policies for extrapolating to a wide range of test conditions with dynamics shifts. In this work, we aim to train diverse policies under the regularization of the behavior patterns. We motivate our paradigm by observing the inverse dynamics in the environment with partial state information and propose \textit{Diversity in Regulation}~(DiR) training diverse policies with regulated behaviors to discover desired patterns that benefit the generalization. Considerable empirical results on various variations of different environments indicate that our method attains improvements over other diversity-driven counterparts.
\end{abstract}

\section{Introduction}
Deep Reinforcement Learning has exhibited wide success in solving complex tasks, including vision-based video games~\cite{mnih2015human,jaderberg2019human}, quadruped locomotion~\cite{hwangbo2019learning,lee2020learning}, and robotic manipulation~\cite{andrychowicz2020learning}. However, the policies trained in the source environments are prone to fail in environmental variations. For example, dynamics change, such as the damaged component of a robot or encountering a new terrain, might lead to a failure due to poor generalization of the agents. Thus one significant challenge for real-world deployment of RL is the generalization across various conditions. 

One natural approach is to train a policy under a range of dynamics in simulation~\cite{tobin2017domain,peng2018sim,rajeswaran2016epopt}, and assume that the trained policy can generalize to the specific target environment. These methods require expert knowledge to manually specify the distribution of training environments in a trial-and-error manner to involve the properties of the target environment. In addition, the trained policy may appear to be over-conservative due to the uncertainty in the training environments~\cite{yu2018policy,xie2021dynamics}. Another category of methods trains the policy to implicitly identify the dynamics of the target environment based on samples collected in the target environment during training and then encourages the policy to act optimally according to the identified dynamics~\cite{muratore2021data,du2021auto,evans2022context}. However, rollouts in the target environment, like the real-world system during training, might result in catastrophic damage due to the premature behavior of the policy.

Recent works have demonstrated that diversity-driven policies can extrapolate to new environments through the few-shot adaptation~\cite{eysenbach2018diversity,kumar2020one,osa2021discovering, parker2020effective,zhou2022continuously}. While the policy population with different behavior characteristics can generalize to different environment variations, the learned policies may result in potential safety problems in practical scenarios like real-world systems, as the behaviors of the diverse policies are unpredictable. Especially, the degree of diversity that is necessary for the generalization may be limited. For instance, when we aim to obtain multiple policies for a quadruped robot that can generalize to various terrains, what we desire might be policies with different locomotion patterns rather than those able to roll on the ground. However, these works train the policies without regularizing the behavior, which might result in inadequate feasible policies.

In this work, we take the first step towards diverse policies with regulated behaviors for generalization. To encourage sufficient feasible solutions for adaptation in a wide range of downstream scenarios, we propose a novel diversity objective based on the divergence of inverse dynamics models $\mathcal{T}^{\pi}(a|f(s),f(s'))$ under partial state information. The partial state information is removed by utilizing a customizable state filtration function $f(s)$. Intuitively, the actions impacting the remaining state information would be discouraged from getting diversified, thus regulating the behaviors of trained policies. Additionally, we introduce the open-ended training manner to achieve continuous solution discovery, which avoids the drawback of prior work training with a fixed number of policies~\cite{kumar2020one,parker2020effective}. 

The main contribution of our work is the proposal of a diversity-driven algorithm, \textit{Diversity in Regulation}~(DiR), which trains multiple policies with regulated behavior patterns for efficient generalization by diversifying the action distributions in a customizable way. Our analysis demonstrates that the discovered policies show more regulated behaviors against prior diversity-driven approaches, which benefits generalization across a wide range of test conditions. Empirically, we observe that DiR substantially outperforms prior methods under various environment discrepancies.

\section{Preliminaries}

\textbf{Notation.}~To model the sequential decision problem, we consider the standard Markov Decision Process~(MDP)~\cite{sutton2018reinforcement} as $(\mathcal{S},\mathcal{A},\mathcal{P},r,\mu,\gamma)$, where $\mathcal{S}$ and $\mathcal{A}$ are the state space and action space respectively; $\mathcal{P}(s'|s,a):\mathcal{S}\times\mathcal{A}\times\mathcal{S}\to[0,1]$ specifies the dynamics of the environment and defines the transition probability of reaching $s'$ at the next step given current state $s$ and the executed action $a$; $r(s,a): \mathcal{S}\times\mathcal{A}\to \mathbb{R}$ denotes the reward function; $\mu(s): \mathcal{S}\to[0,1]$ denotes the distribution of initial states and $\gamma\in(0,1)$ is the discount factor. Considering a policy $\pi(a|s):\mathcal{S}\times\mathcal{A}\to[0,1]$ which outputs the probability of choosing action $a$ given the state $s$, the probability density function of any trajectory $\tau=\{s_0,a_0,s_1,a_1,\dots,s_T\}$ can be formulated as $\mathrm{P}(\tau)=\mu(s_0)\prod_{t=0}^{T-1} \pi(a_t|s_t)\mathcal{P}(s_{t+1}|s_t,a_t)$.

\textbf{Inverse dynamics.} Here we denote the inverse dynamics of the MDP as $\mathcal{T}(a|s,s'):\mathcal{S}\times\mathcal{S}\times\mathcal{A}\to[0,1]$ which defines the probability of action $a$ given the state pair $(s,s')$ in the consecutive steps. Since there might be various actions under different policies given the state pair $(s,s')$, the inverse dynamics under some specific policy $\pi$ can be formulated as:
\begin{equation}
    \mathcal{T}^\pi(a|s,s')=\dfrac{\mathcal{P}(s'|s,a)\pi(a|s)}{\int_\mathcal{A}\mathcal{P}(s'|s,\hat{a})\pi(\hat{a}|s)d\hat{a}}~.
    \label{eq:id}
\end{equation}

\textbf{State filtration function.} In this work, we assume a customizable function $f(s):\mathcal{S}\to\bar{\mathcal{S}}$ that removes partial state information from the state s. For instance, $f(s)$ can be defined to remove the x-axis coordinate in a navigation task whose full state information includes 2D coordinates. The resulting partial state space $\bar{\mathcal{S}}$ can also be considered as the state space from a partially observable MDP~(POMDP)~\cite{bellman1957markovian}.

\textbf{Mutual-Information in RL.}~Mutual information can be generally expressed as 
\begin{align}
    &I(X;Y) = \iint p(x,y)\log\dfrac{p(x,y)}{p(x)p(y)}dydx\nonumber\\
    &~\quad\qquad = H(X)-H(X|Y) = H(Y)-H(Y|X),
\end{align}
which defines the mutual dependence between two random variables. Mutual information has been introduced to find the best representation subject to a constraint on the complexity~\cite{tishby2000information,alemi2016deep}. In the context of RL, maximizing the mutual information $I(S;Z)=H(Z)-H(Z|S)$ between visited states $S$ and latent variable $Z$ has been proposed to discover diverse policies~\cite{eysenbach2018diversity,kumar2020one}. 

\textbf{Diverse high-performing policies.}~
Prior approaches that train multiple high-performing policies $\{\pi_{\theta_k}\}_{k=1}^M$ with diverse behaviors\footnote{The $\theta$ will be omitted for simplicity.} can be naturally formulated as:
\begin{align}
    &\max_{\pi_k}~\mathbb{E}_{\pi_{k}}\left[
    \sum_{t=1}^T \gamma^{t-1}r_t
    \right],~\forall~1\leq k \leq M \label{eq:diverse basic}\\
    &\text{s.t.}~D\left(\pi_{i}, \pi_{j}\right)\geq\delta,~\forall~1\leq i,j\leq M,~i\neq j\nonumber,
\end{align}
where $D(\cdot,\cdot)$ measures how different the two policies are, and $\delta$ is the diversity threshold. One natural choice of the distance measurement function is the KL divergence $\mathrm{D}_{KL}\left[\pi_{i}(\cdot|s)\parallel\pi_{j}(\cdot|s)\right]$ which is widely adopted in prior works~\cite{schulman2015trust,schulman2017proximal,hong2018diversity}.

\textbf{Few-shot adaptation.}~
In novel environments, we rollout each policy from the trained population for several episodes and deploy the best-performing one, which resembles prior diversity-driven approaches~\cite{kumar2020one, osa2021discovering, derek2021adaptable}.

\begin{figure}[t]
    \centering
    % \footnotesize
    \includegraphics[width=0.47\textwidth]{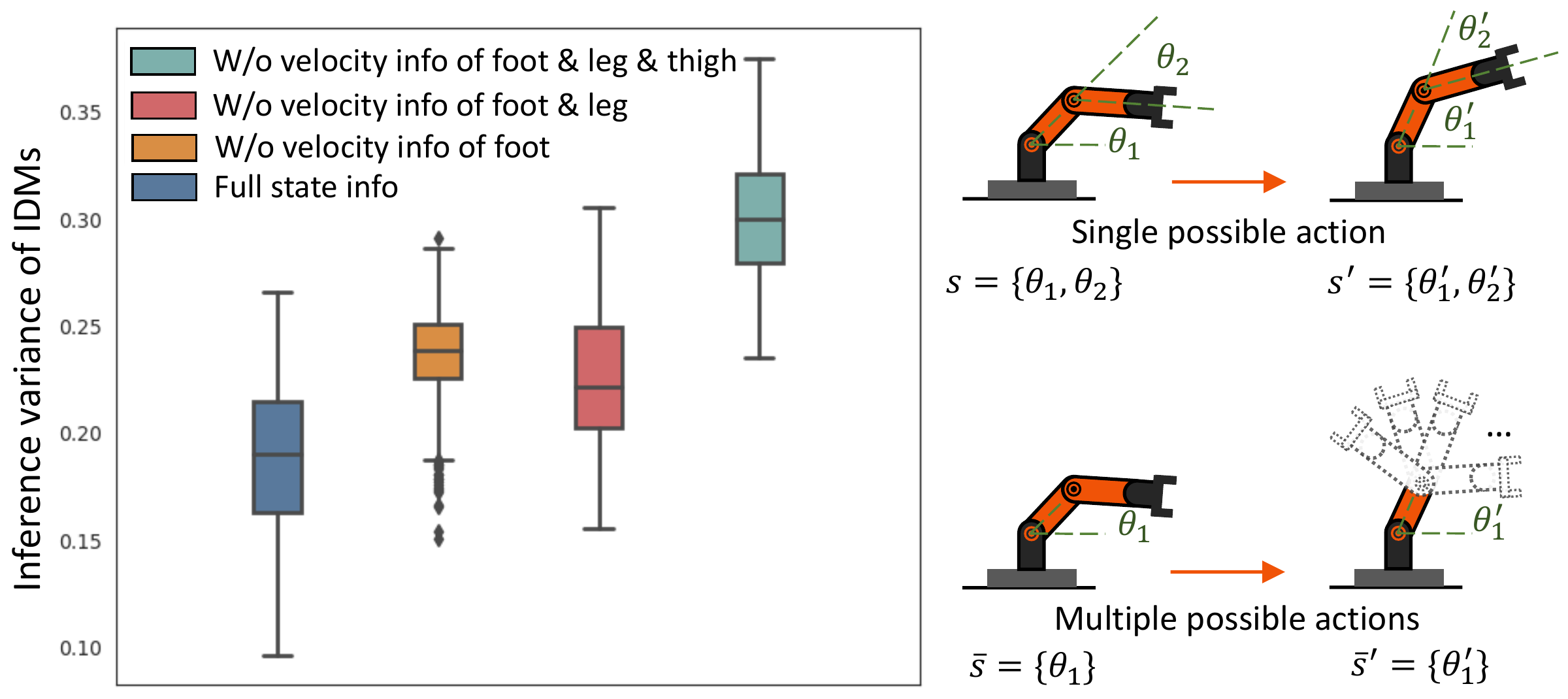}
    % \vspace{-0.5em}
    \caption{Left: Inference variance of inverse dynamics models trained under different state settings in Walker2D increases with the increase of missing state information. Right: Toy example of a 2-DOF robotic arm to interpret the inference variance of IDMs under different state settings. Given the partial state information pair $(\bar{s},\bar{s}')$, there will be more possible actions compared with the full state information setting.}
    % \vspace{-0.5em}
    \label{fig:motivation}
\end{figure}

\section{Motivation Example: Inference Variance of Inverse Dynamics Models}
We motivate our method with an empirical observation about the inference variance of the inverse dynamics under different state information settings. Here we train four independent inverse dynamics models~(IDMs) $\{\mathcal{T}_{\phi_i}(a|s,s')\}_{i=1}^4$ simultaneously using different state settings in Walker from Mujoco~\cite{todorov2012mujoco}. The training data are collected by an agent trained through Soft Actor-Critic~(SAC)~\cite{haarnoja2018soft}. We utilized four different state settings for the inverse dynamics models with varying degrees of information missing. We train the inverse dynamics models by maximizing the log-likelihood:
\begin{equation}
    L(\phi_i) = \mathbb{E}_{(s,a,s')\sim \mathcal{D}}\left[\log
     \mathcal{T}_{\phi_i}\left(a| f_i(s),f_i(s')\right)
    \right],~1\leq i\leq4,
\end{equation}
where the $f_i(s):\mathcal{S}\to\bar{\mathcal{S}_i}$ is the state filtration function corresponding to the model $i$ that removes some specific state information. After training, we evaluate the inference variance $\mathrm{Var}\left[\mathcal{T}_{\phi_i}(\cdot|s,s')\right]$ of the IDMs given the same batch of data $\mathcal{D}_{test}=\{(s,s')\}$, and the results are shown in Figure~\ref{fig:motivation}. The results indicate that the inference variance of the inverse dynamics model increases with the increase of missing state information. The details of the experiment refer to Appendix~A.%\ref{apd:motivation}.

Herein we interpret the observation with a toy example of a 2-DOF robotic arm, as shown in Figure~\ref{fig:motivation}. When we employ the joint angle $\theta_1$ as a partial state space $\bar{\mathcal{S}}$, there might be multiple possible actions given any state pair $(\bar{s},\bar{s}')$. In contrast, there might be only one possible action given the state pair $(s,s')$ under the full state information setting. Motivated by the observation, we propose to maximize the divergence of IDMs with the partial state setting under different policies to encourage them to produce distinct action distributions given any $(f(s),f(s'))$ pair. Furthermore, we introduce customizable state filtration functions to specify the desired patterns. Semantically, the removing state information~(e.g.,~$\theta_2$ above) can be regarded as the state information of body parts that is unnecessary to be diversified~(e.g.,~roll angles of a quadruped robot). In contrast, we can remove the state information like leg motions to obtain policies with diverse locomotion patterns. Thus, we introduce the state filtration function to focus on discovering desired behaviors.

We denote the divergence between inverse dynamics under the partial state information setting as $\mathrm{P}^{\pi_j}(a|f(s),f(s')):\bar{\mathcal{S}}\times\bar{\mathcal{S}}\times\mathcal{A}\to[0,1]$, and the overall objective can be formulated as the following constrained optimization problem:
\begin{align}
    &\qquad\quad\max_{\pi_k}~\mathbb{E}_{\pi_k}\left[
    \sum_{t=1}^T \gamma^{t-1}r_t
    \right],~\forall~1\leq k \leq M \label{eq:main} \\
    &\text{s.t.}~\mathbb{E}\left[\mathrm{D}_{KL}\left[\mathrm{P}^{\pi_i}(\cdot|f(s),f(s'))\parallel\mathrm{P}^{\pi_j}(\cdot|f(s),f(s'))\right]\right]
    \geq\delta,\nonumber\\
    &\qquad\qquad\qquad\quad\forall~1\leq i,j\leq M,~i\neq j\nonumber,
\end{align}
where $f(s):\mathcal{S}\to\bar{\mathcal{S}}$ removes specific partial state information from $s$, and can be designed to control the diverse patterns we aim to discover.

\section{Diversity in Regulation}
\begin{figure}[t]
    \centering
    % \footnotesize
    \includegraphics[width=0.45\textwidth]{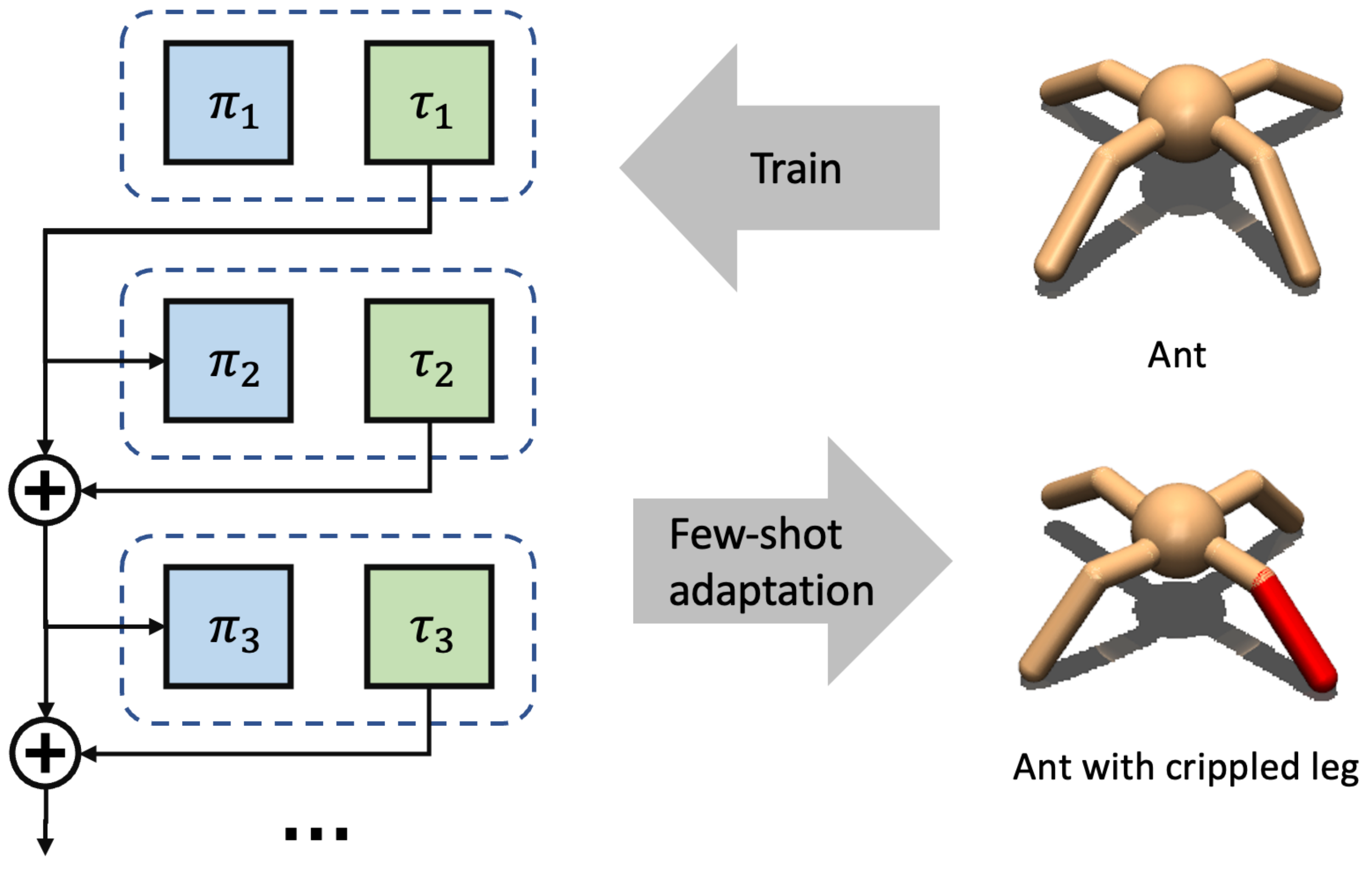}
    \caption{The semantic overview of DiR. $\pi_i$ and $\tau_i$ represent the policy and inverse dynamics model, respectively. We train multiple policies in the source environments in an iterative manner and generalize to the variation condition with the best-performing one after the few-shot adaptation.}
    % \vspace{-1.5em}
    \label{fig:semantic}
\end{figure}

This section presents our approach to resolving the objective in detail. We propose an open-ended training manner for diverse solution discovery, transform the objective into a trivial form, and finally analyze the connection between our method and prior diversity-driven approaches through mutual information. The overview of DiR is shown in Figure.~\ref{fig:semantic}.

\subsection{Open-ended Solution Discovery}
Directly solving Eq.~\ref{eq:diverse basic} or Eq.~\ref{eq:main} requires the non-trivial pairwise constraints computation for training each policy and the parallel framework for the policy population, which limits the population size in prior works~\cite{parker2020effective,masood2019diversity}. Here we introduce an iterative training manner that trains only a single policy at a time and optimizes the policy $\pi_k$ to be distinct from the previously discovered policies $\{\pi_i\}_{i=1}^{k-1}$, which resembles the prior work~\cite{zhou2022continuously}. However, we concentrate on the regulated diversity objective. Formally, we optimize the following objective to train a policy $\pi_k$ at each iteration:
\begin{align}
    &\qquad\qquad\qquad\quad\max_{\pi_k}~\mathbb{E}_{\pi_k}\left[
    \sum_{t=1}^T \gamma^{t-1}r_t
    \right] \label{eq:open-ended} \\
    &\text{s.t.}~\mathbb{E}_{\pi_k}\left[\mathrm{D}_{KL}\left[\mathrm{P}^{\pi_k}(\cdot|f(s),f(s'))\parallel\mathrm{P}^{\pi_i}(\cdot|f(s),f(s'))\right]\right]\geq\delta,\nonumber\\
    &\quad\qquad\qquad\qquad\qquad\quad\forall~1\leq i< k\nonumber,
\end{align}
which converts the optimization of the whole population simultaneously to the optimization of a single policy with a simplified constraint. Furthermore, we can discover policies exhaustively through open-ended training to obtain some specific behavior pattern we desire, which is superior to prior works that fix the number of policies.

\subsection{Diversity via Inverse Dynamics Disagreement}
To solve the constrained optimization problem in Eq.~\ref{eq:open-ended}, we introduce the Lagrangian multiplier method to convert the hard constraints to soft penalties:
\begin{align}
    &\quad~\max_{\pi_k}~\mathbb{E}_{\pi_k}\left[
    \sum_{t=1}^T \gamma^{t-1}r_t\right] + \sum_{i=1}^{k-1}\beta_i Div(\pi_k,\pi_i), \label{eq:lagrangian} \\ 
    &\text{where}~Div(\pi_k,\pi_i)=\nonumber\\
    &\quad\mathbb{E}_{\pi_k}\left[\mathrm{D}_{KL}\left[\mathrm{P}^{\pi_k}(\cdot|f(s),f(s'))\parallel\mathrm{P}^{\pi_i}(\cdot|f(s),f(s'))\right]\right], \nonumber
\end{align}
where $\{\beta_i\}_{i=1}^{k-1}$ are the multipliers that can be considered as hyperparameters, and $Div(\pi_k,\pi_i)$ can be interpreted as the inverse dynamics disagreement between two policies. Introducing the Lagrangian multipliers method to simplify the constrained optimization problem and set the multipliers as hyperparameters is widely used in the RL research~\cite{stooke2020responsive,chane2021goal,peng2018sim}. While the constraints might exhibit oscillations during training~\cite{stooke2020responsive}, it is acceptable since we aim to encourage diversity rather than obtain severely distinct policies.

To tractably optimize the $Div(\pi_k,\pi_i)$, we introduce an ensemble of inverse dynamics models $\{\mathcal{T}_{\phi_i}\}_{i=1}^M$ to approximate the inverse dynamics under corresponding policies. At each iteration, we train the inverse dynamics model $\mathcal{T}_{\phi_k}$ simultaneously by maximizing the log-likelihood:
\begin{equation}
    L(\phi_k)=\mathbb{E}_{\pi_k}\left[\log\mathcal{T}_{\phi_k}(a|f(s),f(s'))\right].
\end{equation}
To solve the optimization for policy $\pi_k$ in Eq.~\ref{eq:lagrangian}, we approximate the $\mathrm{P}^{\pi_i}(a|f(s),f(s'))$ with the an inverse dynamics model $\mathcal{T}_{\phi_i}(a|f(s),f(s'))$, and thus convert the diversity objective as follows:
\begin{align}
    Div_{ce}(\pi_k,\pi_i)=\mathbb{E}_{\pi_k}\left[
    -\log \mathcal{T}_{\phi_i}(a|f(s),f(s'))
    \right],\label{eq:new div}
\end{align}
which can be interpreted as the cross-entropy between the transitions collected by policy $\pi_k$ and the inverse dynamics of $\pi_i$. We present in Appendix~B.1 that the diversity objective in Eq.~\ref{eq:new div} can be approximately lower bounded by the objective in Eq.~\ref{eq:lagrangian} with less penalty to the entropy of the policy.
By introducing the novel diversity objective, we present the final objective that optimizes diversity through the transformed inverse dynamics disagreement:
\begin{align}
    &J(\theta_k)=\mathbb{E}_{\pi_k}\left[
    \sum_{t=1}^T \gamma^{t-1}r_t\right] \label{eq:final}+\dfrac{\alpha}{k-1} \sum_{i=1}^{k-1}Div_{ce}(\pi_k,\pi_i),
\end{align}
where $\alpha$ is a scaling hyperparameter. Since we aim to optimize the policy $\pi_k$ to be distinct from each previously policies without any preference, we set the multipliers $\{\beta\}_{i=1}^{k-1}$ from Eq.~\ref{eq:lagrangian} as $\frac{1}{k-1}$.

For implementation, we convert the diversity objective in Eq.~\ref{eq:final} to an intrinsic reward, which trivially optimizes the objective. Herein, we define the DiR reward function as:
\begin{equation}
    r^{DiR}_t = r_t + \dfrac{\alpha}{k-1}\sum_{i=1}^{k-1}\left[-\log \mathcal{T}_{\phi_i}(a_t|f(s_t),f(s_{t+1}))\right].
\end{equation}
We implement DiR with Proximal Policy Optimization~(PPO)~\cite{schulman2017proximal}, and we train an ensemble of policies $\{\pi_{\theta_k}\}_{k=1}^M$ and IDMs $\{\mathcal{T}_{\phi_i}\}_{i=1}^M$ iteratively. Note that we only adopt the state filtration function for the IDMs rather than the policies. The pseudo-codes of DiR and the few-shot adaptation can be found in 
% Alg.~\ref{alg:dir} and Alg.~\ref{alg:adap} from 
Appendix~B.2.
% , respectively.

\begin{figure*}[t]
    \centering
    % \footnotesize
    % \vspace{-1em}
    \includegraphics[width=\textwidth]{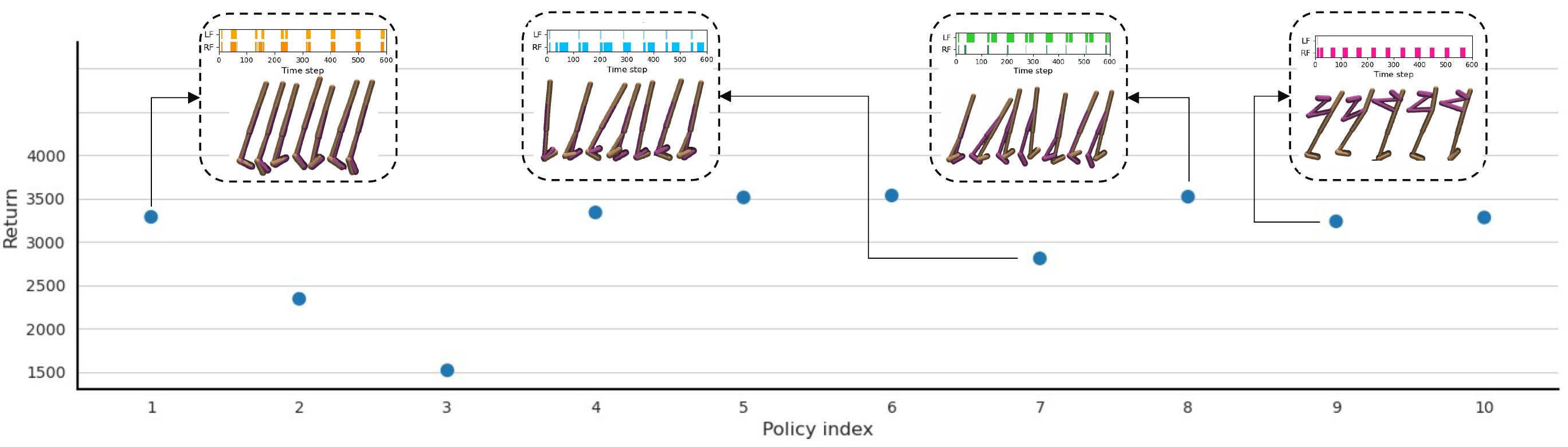}
    \caption{Visualization of the discovered policies in Walker. The y-axis represents the performance of discovered policies during the iterative training. We show foot contact patterns and motion illustrations of four policies in the dotted boxes. The shaded areas mark the time steps during which the respective foot (LF: left foot or RF: right foot) is in contact with the ground.}
    % \vspace{-1.5em}
    \label{fig:diverse traj}
\end{figure*}

\subsection{Connections to Prior Work}
Here we denote different policies from a policy ensemble as a random variable $z$. Several diversity-driven approaches maximizing the divergence between the inference action distributions of different policies on expectation~\cite{parker2020effective,derek2021adaptable} can be formulated as maximizing $I(a;z|s)=H(a|s)-H(a|s,z)$. Additionally, the unsupervised skill discovery works focus on diversifying the state occupancy by maximizing $I(s;z)=H(s)-H(s|z)$~\cite{eysenbach2018diversity} or $I(s';z|s)=H(s'|s)-H(s'|s,z)$~\cite{Sharma2020Dynamics-Aware}. 

Similarly, our proposed DiR diversity objective can also be interpreted as a conditional mutual information
\begin{equation}
    I(a;z|\bar{s},\bar{s}')=H(a|\bar{s},\bar{s}')-H(a|\bar{s},\bar{s}',z),
    \label{eq:mutual}
\end{equation} 
where $\bar{s}:=f(s)$ and $\bar{s}':=f(s')$. Intuitively, we encourage the output actions of different policies to be discriminable given the same partial state pair $(\bar{s},\bar{s}')$. The state filtration function controls the degree of diversity. When the state filtration function is the identity function such that $f(s)=s$, there will be no further diversity as the action is relatively certain given the full state pair $(s,s')$. In contrast, the policies will be optimized to e xexecute distinct actions at all time if $f(s)=\O$ removes all state information. Thus, we can regulate the diverse behaviors to a specific coverage of patterns by customizing the state filtration function.

\section{Experiment}
In this section, we aim to empirically answer the following questions: (1) Can our method discover diverse behaviors? (2) Does our method discover diverse policies with regulated behavior patterns through the state filtration function? (3) Since we hypothesize DiR can obtain more feasible policies compared with baselines given the same population size, can the trained population perform better in a wide range of dynamics mismatch scenarios? Implementation details and additional results are presented in Appendix.

\subsection{Experimental Settings}
\textbf{Environments.}~
We adopt four continuous control tasks: Ant, Walker, Hopper, and Minitaur from Mujoco~\cite{todorov2012mujoco} and Bullet~\cite{coumans2019}, as illustrated
% in Figure~\ref{fig:components} 
in Appendix.~C. We implement extensive test scenarios, including broken leg joints, shifted dynamics parameters, and sensor failure conditions. See Appendix~C.1 for details of the environments.

\noindent\textbf{Customized state filtration functions.}~
We focus on discovering diverse locomotion patterns~(e.g., walking with different legs in Ant) by designing the state filtration function for the four locomotion tasks. Thus we remove partial state information about the leg motions~(e.g., joint positions) through $f(s)$ in all four environments. 
%In Ant, $f(s)$ removes all state information of the four legs. In Hopper, $f(s)$ removes the state information of the leg. In Walker, $f(s)$ removes the state information of the right leg. In Minitaur, $f(s)$ removes the joint positions of all motors. 
Full details of the state filtration functions and the original state information are described in Appendix~C.2.

\noindent\textbf{Baselines and implementation.}~
We compare DiR to SMERL~\cite{kumar2020one} that trains diverse policies for generalization to environmental variations, DvD~\cite{parker2020effective} that trains an ensemble of policies via the proposed divergence determinant, vanilla PPO with multiple independent policies~(Multi), vanilla PPO with a single policy~(PG). 
%The comparison with SMERL and DvD is that these two baselines maximize the diversity through the state occupancy and action occupancy, which is different from our proposed conditional action occupancy as shown in Eq.~\ref{eq:mutual}. 
We set the population size as $10$ in all baselines, and all baselines except PG train the same number of policies. We train each policy with $2$M steps for all algorithms. Each trial runs eight times with different random seeds. Refer to Appendix C.3 for implementation details.

\subsection{Emergent Behavior with DiR}

Since we remove the state information of one leg in Walker, we hypothesize that our method can learn various locomotion patterns (e.g.,~hopping, walking) with different legs. Thus, we visualize the foot contact within an episode in Walker, as shown in Figure~\ref{fig:diverse traj}. The results suggest that our method can iteratively discover diverse policies with different locomotion patterns, including hopping on both feet, incomplete one-leg hopping, and complete one-leg hopping. Furthermore, there is no significant performance degradation resulting from diversity-driven training.

\subsection{Population Comparison in Training Environments}
Herein we aim to compare the policies with prior diversity-driven approaches quantitatively. We adopt the population diversity, a determinant-based diversity paradigm proposed in~\cite{parker2020effective}, to quantify the behavior diversity of the trained policies. Here we collect 2000 states for the behavior embeddings in each environment. The results are reported in Table~\ref{tab:diversity score}, where DiR outperforms all baseline methods concerning the diversity score. We observe that the independently trained policies in Multi can also obtain competitive diversity scores compared to DvD and SMERL which introduce extra diversity objectives, which may result from the different initialization of the policies as presented in~\cite{jiang2021emergence}.

\begin{figure}[H]
    % \vspace{-0.5em}
    \centering
    % \subfloat[\centering DiR]{{\includegraphics[width=0.115\textwidth]{figure/dir_feet_contact.pdf} }}%
    % %\quad
    % \subfloat[\centering SMERL]{{\includegraphics[width=0.115\textwidth]{figure/smerl_feet_contact.pdf} }}%
    % %\quad
    % \subfloat[\centering DvD]{{\includegraphics[width=0.115\textwidth]{figure/dvd_feet_contact.pdf} }}%
    % %\quad
    % \subfloat[\centering Multi]{{\includegraphics[width=0.115\textwidth]{figure/multi_feet_contact.pdf} }}%
    \includegraphics[width=0.47\textwidth]{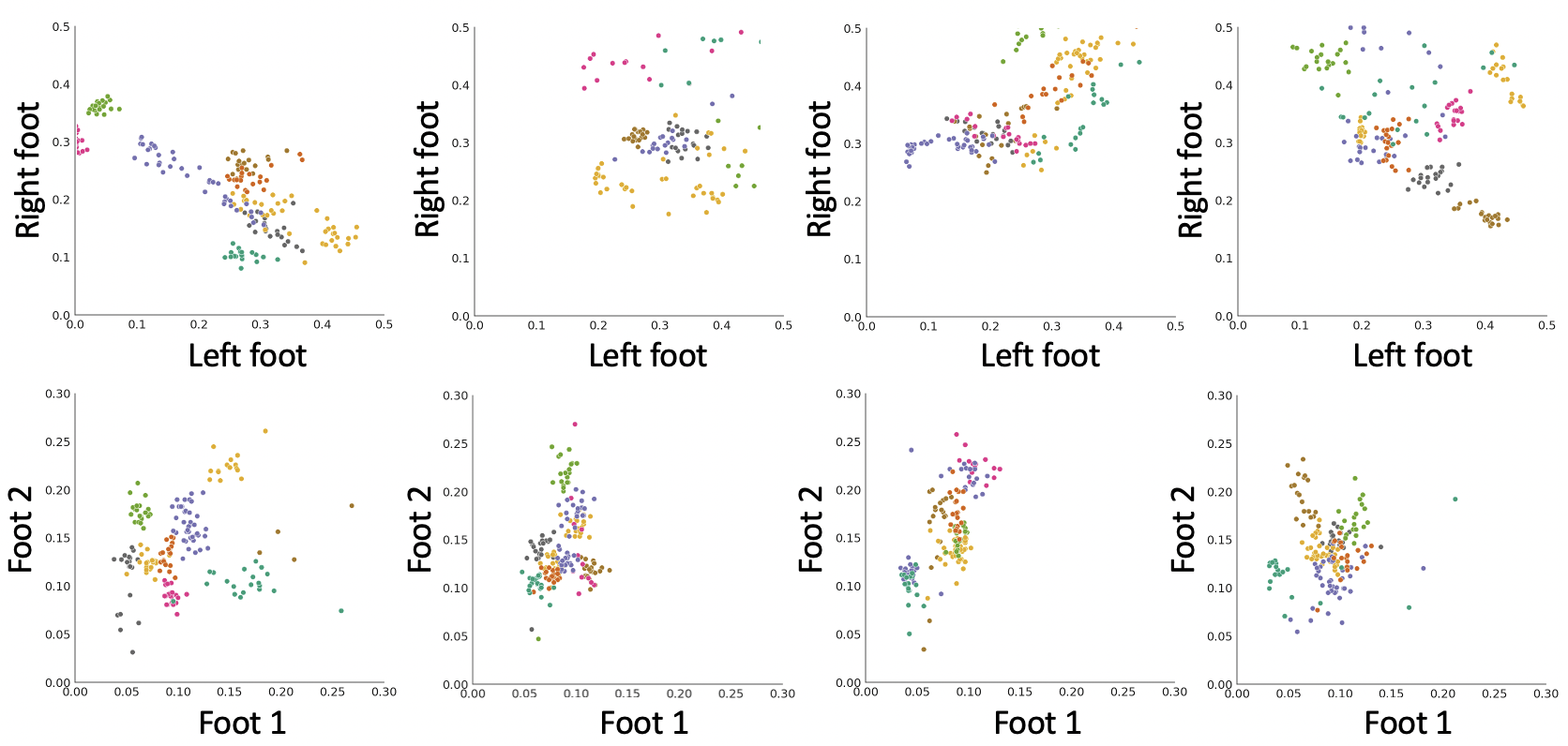}
    % \vspace{-0.5em}
    % \footnotesize
    \caption{Visualization of behaviors through the time proportion when feet contact the ground, where the different colors indicate different policies. Different colors represent different policies from the population, and we run each policy for 20 episodes. (Top): Time proportion when two feet contact the ground within an episode in Walker. (Bottom): Time proportion when two adjacent feet contact the ground within an episode in Ant.}
    % \vspace{-1em}
    \label{fig:contact ground}%
\end{figure}

To examine whether DiR can train more practical locomotion patterns by inducing the regulated diversity objective, we adopt the approach that describes behaviors of quadruped robots in prior works~\cite{cully2015robots,nilsson2021policy}, which computes the time proportion when the feet contact the ground within an episode. The visualization of the behavior diversity in Walker and Ant is shown in Figure~\ref{fig:contact ground}. The results show that the behaviors of each policy trained by DiR are more consistent in multiple episodes, and DiR can discover more distinct locomotion patterns compared with baselines.

\begin{figure}[h]
    \centering
    % \footnotesize
    \includegraphics[width=0.45\textwidth]{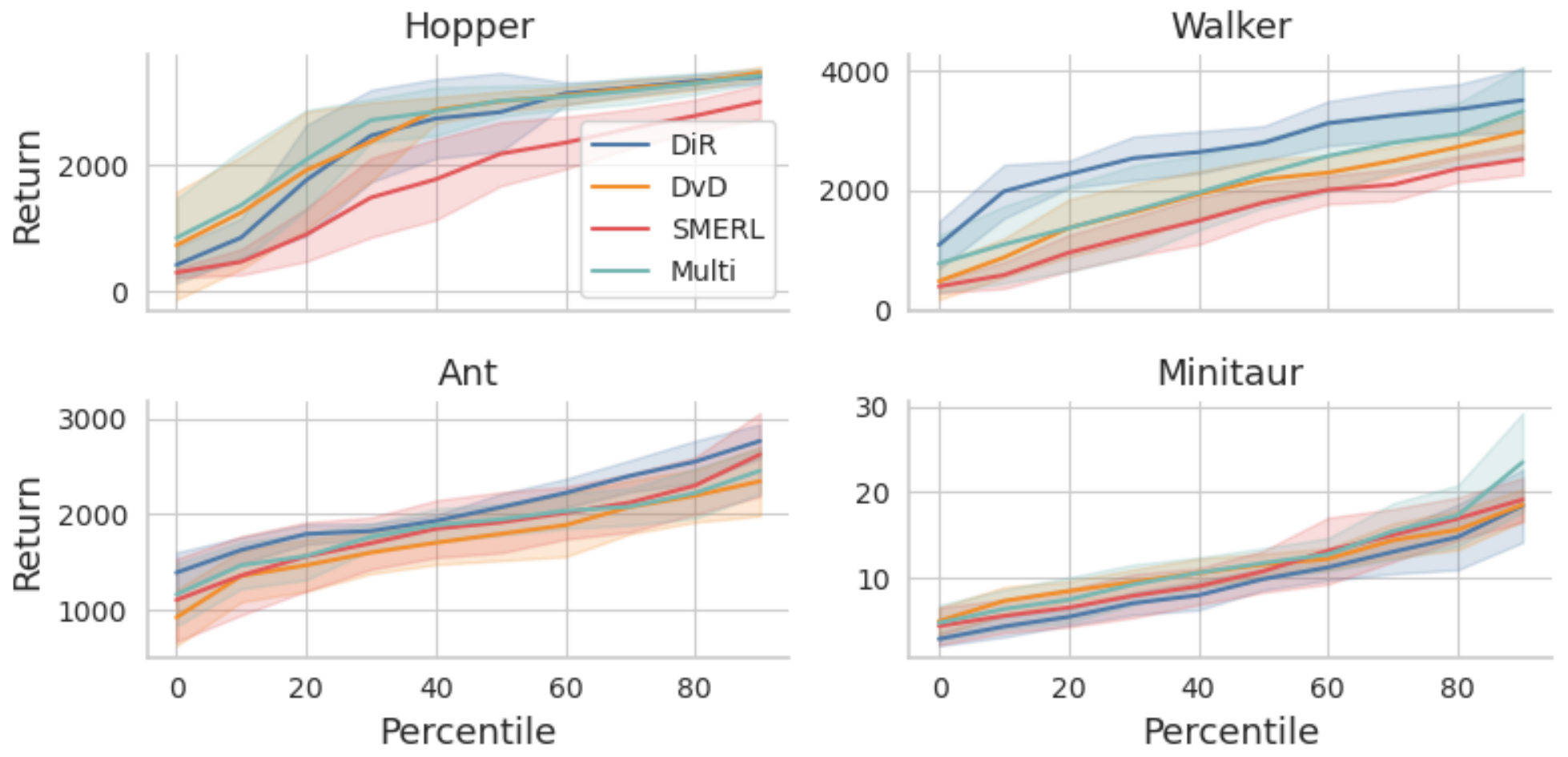}
    \captionof{figure}{Percentile performance of the policy population across all four environments. The 0 percentile on the x-axis represents the worst-performing policy, while the 100 percentile represents the best-performing one.}
    \label{fig:performance}
\end{figure}

Furthermore, here we show the percentile performance of the population, as shown in Figure~\ref{fig:performance}. The results indicate that DiR achieves competitive performance in Hopper and Minitaur while obtaining better averaged and worst-case performance in Walker and Ant, compared with baselines. We believe that regulated diversity can hinder the discovery of behavior patterns with poor performance, which results in improved worst-case performance over the policies.

% \begin{table}
%     % \footnotesize
%     % \scriptsize
%     \captionof{table}{Diversity scores across all environments. Asterisks indicate that the results are significantly different from all the baselines under $\mathrm{p} < 0.05$.}
%     \label{tab:diversity score}
%     \begin{tabular}{ccccc}
%         \toprule  
%         ~ & Ant & Hopper & Walker & Minitaur   \\  
%         \midrule
%         Multi & $71.69(0.40)$ & $57.88(2.46)$ & $76.46(0.42)$ & $81.64(0.16)$ \\
%         \midrule
%         DvD & $71.59(0.49)$ & $59.08(2.20)$ & $76.24(0.38)$ & $81.43(0.20)$ \\
%         \midrule
%         SMERL & $71.47(0.53)$ & $63.18(2.39)$ & $77.10(0.40)$ & $81.73(0.26)$ \\
%         \midrule
%         DiR & $\textbf{73.01(0.35)}^*$ & $\textbf{66.07(1.69)}^*$ & $\textbf{78.36(0.83)}^*$ & $\textbf{83.23(0.27)}^*$ \\
%         \bottomrule
%     \end{tabular}
% \end{table}

\begin{table}
    \footnotesize
    % \scriptsize
    \begin{tabular}{ccccc}
        \toprule  
        ~ & Ant & Hopper & Walker & Minitaur   \\  
        \midrule
        Multi & $71.6(0.4)$ & $57.8(2.4)$ & $76.4(0.4)$ & $81.6(0.2)$ \\
        \midrule
        DvD & $71.6(0.5)$ & $59.1(2.2)$ & $76.2(0.4)$ & $81.4(0.2)$ \\
        \midrule
        SMERL & $71.5(0.5)$ & $63.2(2.4)$ & $77.1(0.4)$ & $81.7(0.3)$ \\
        \midrule
        DiR & $\textbf{73.0(0.4)}^*$ & $\textbf{66.1(1.7)}^*$ & $\textbf{78.4(0.8)}^*$ & $\textbf{83.2(0.3)}^*$ \\
        \bottomrule
    \end{tabular}
    \captionof{table}{Diversity scores across all environments. Asterisks indicate that the results are significantly different from all the baselines under $\mathrm{p} < 0.05$.}
    \label{tab:diversity score}
    % \vspace{-1em}
\end{table}

% \begin{figure*}[t]
%     %\centering
%     \begin{minipage}{0.46\textwidth}\centering
%         \includegraphics[width=\textwidth]{figure/Return_percentile_Distribution_for_all_env.pdf}
%         \vspace{-1em}
%         \footnotesize
%         \captionof{figure}{Percentile performance of the policy population across all four environments. The 0 percentile on the x-axis represents the worst-performing policy, while the 100 percentile represents the best-performing one.}
%         \label{fig:performance}
%     \end{minipage}
%     \qquad\quad
%     \begin{minipage}{.47\textwidth}\centering
%         % \footnotesize
%         \scriptsize
%         \captionof{table}{Diversity scores across all environments. Asterisks indicate that the results are significantly different from all the baselines under $\mathrm{p} < 0.05$.}
%         \label{tab:diversity score}
%         \begin{tabular}{ccccc}
%             \toprule  
%             ~ & Ant & Hopper & Walker & Minitaur   \\  
%             \midrule
%             Multi & $71.69(0.40)$ & $57.88(2.46)$ & $76.46(0.42)$ & $81.64(0.16)$ \\
%             \midrule
%             DvD & $71.59(0.49)$ & $59.08(2.20)$ & $76.24(0.38)$ & $81.43(0.20)$ \\
%             \midrule
%             SMERL & $71.47(0.53)$ & $63.18(2.39)$ & $77.10(0.40)$ & $81.73(0.26)$ \\
%             \midrule
%             DiR & $\textbf{73.01(0.35)}^*$ & $\textbf{66.07(1.69)}^*$ & $\textbf{78.36(0.83)}^*$ & $\textbf{83.23(0.27)}^*$ \\
%             \bottomrule
%         \end{tabular}
%     \end{minipage}

%     \vspace{-1.5em}
% \end{figure*}

\subsection{Adaptation in Environment Variations}
To examine whether DiR can provide adequate feasible policies which benefit the generalization, we implement various variations of the environments, including the crippled legs, the shifts of the dynamics parameters~(e.g., mass), and sensor failures. See Appendix C.1 for detailed descriptions of the test conditions. For few-shot adaptation, we run each policy in the environment for 20 episodes and report the performance of the best-performing one. 

We first evaluate the approaches under conditions with damaged body components. As the results show in Table~\ref{tab:damage}, DiR outperforms baseline methods on most test conditions. In the variations of Hopper, DiR achieves comparable performance with baselines, which can result from the morphology with only one leg that limits the possible locomotion patterns. However, given the other three environments where the robots are multi-legged, DiR surpasses baselines thanks to the regulated diversity discovery. As we utilize the state filtration functions that remove the state information about the legs, DiR will focus on discovering policies that behave differently in terms of the leg motions. Thus, DiR has sufficient strategies~(e.g., walking on a single leg) to handle the situations like the damaged leg. Specifically, DiR is the only approach that adapts with better locomotion patterns in the test Walker environment where a foot joint is broken. Furthermore, we remark that PG training a single policy performs significantly worse than the methods training multiple policies, which indicates that the diversity-driven approaches are simple yet effective for adaptation under dynamics variations. Importantly, we verify that the superior adaptation performance of DiR results from different policies, which further validates that the diverse behaviors benefit the extrapolation to different environments. The details of selected policies in the test environments are presented in 
% Tables~\ref{tab:adap_walker}, \ref{tab:adap_hopper}, \ref{tab:adap_ant} of 
Appendix~D.1.

\begin{figure}[h]
    % \vspace{-1em}
    \centering
    \includegraphics[width=.47\textwidth]{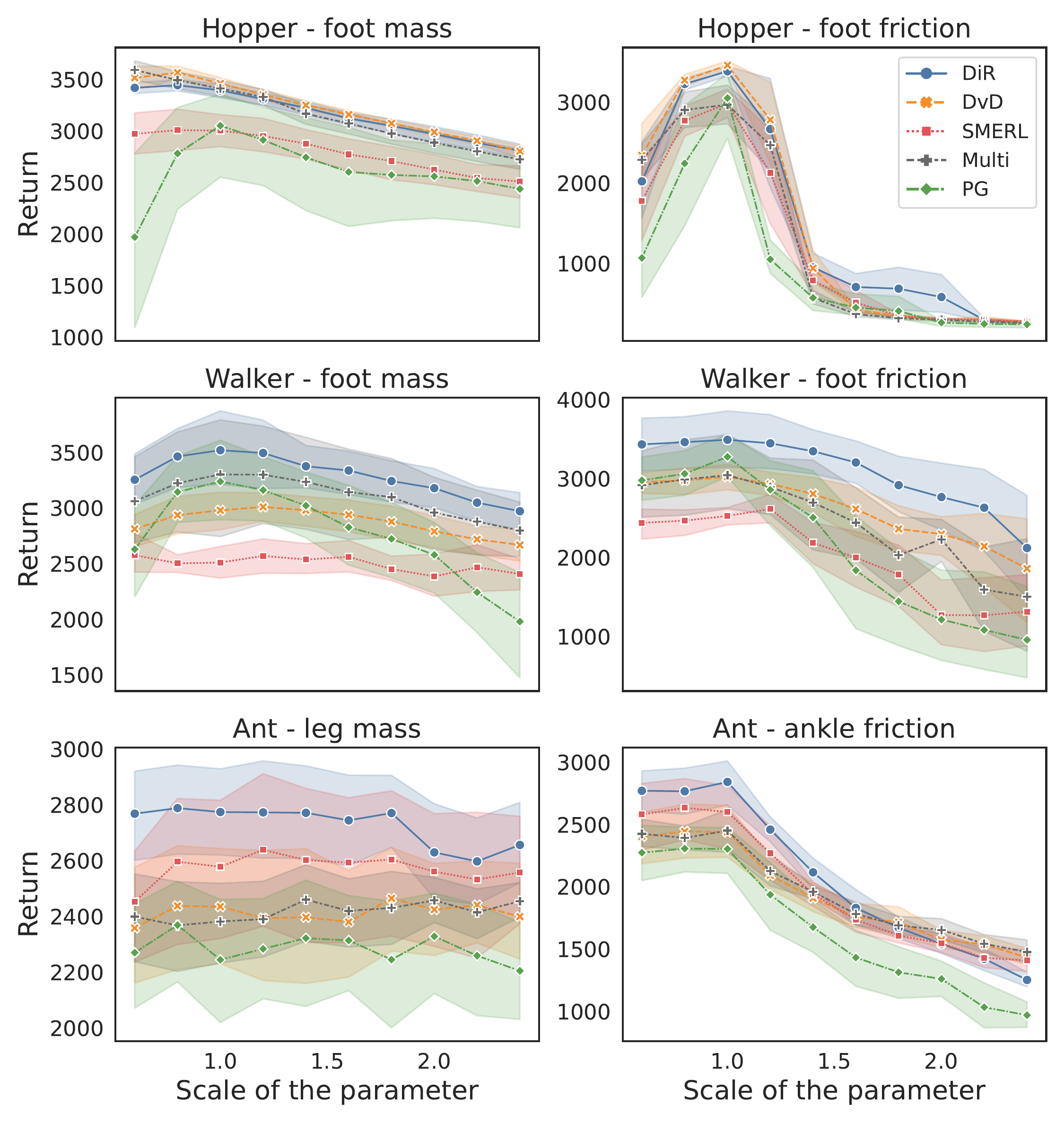}
    % \vspace{-1.5em}
    \caption{Adaptation performance under different levels of the dynamics parameter variations.}
    \label{fig:dynamics}
    % \vspace{-1.8em}
\end{figure}

Furthermore, we consider the adaptation to environments with shifted dynamics parameters. Here we scale the ankle friction or the leg mass of Ant, and the foot friction or the foot mass of Hopper and Walker. As the adaptation performance in Figure~\ref{fig:dynamics} shows, DiR outperforms baselines in Walker and Ant, while DiR produces comparable results with baselines in Hopper. Specifically, DiR provides a non-trivial improvement in \textit{Walker - foot friction} and \textit{Ant - leg mass} compared to all baselines, where the performance does not drop significantly when the value of foot friction gradually increases. Furthermore, we observe that DiR achieves better performance in the environment where the dynamics parameter is the same as the training environment~($\text{scale}=1$), which can result from the stationary diversity-driven intrinsic rewards provided by the fixed inverse dynamics models converged in early iterations. In contrast, DvD and SMERL, whose intrinsic rewards are non-stationary due to the simultaneously trained population or discriminators, might cause unstable training dynamics and thus result in performance degradation. The phenomenon further validates the advantage of the open-ended training manner.

Finally, we implement test environments with various sensor failures where the corresponding state variables are always zeros and report the adaptation performance of the policies. The results in sensor failure conditions of Ant are shown in Table~\ref{tab:sensor failure ant}, where we observe that DiR is weaker than baselines when the sensors on leg 2 are defective, which might result from reliance on the state information of Leg 2 for the decision making. However, DiR outperforms baseline methods in most environments, which validates that the policies trained through DiR output actions conditioning on different state variables of the state information and further verify the robustness of DiR. Additional results in Walker are shown in Appendix~D.2.

\begin{table*}[t!]
    \centering
    % \vspace{-.5em}
    % \footnotesize
    \begin{tabular}{l|rrrr}
        \toprule  
        Environment - Damage & Multi & DvD & SMERL & DiR\\
        \midrule
        Hopper - Broken leg & $2972.0\pm177.0$ & $2798.8\pm838.6$ & $2193.5\pm617.5$ & $2587.6\pm756.7$ \\
        Hopper - Broken foot & $999.5\pm0.2$ & $999.5\pm0.3$ & $999.6\pm0.2$ & $999.3\pm0.4$ \\
        \midrule
        Walker - Broken leg & $2868.1\pm315.2$ & $2677.8\pm268.4$ & $2307.8\pm221.3$ & $\textbf{3059.7}\pm\textbf{199.2}$ \\
        Walker - Broken foot & $1005.9\pm0.5$ & $1015.8\pm15.7$ & $1009.9\pm7.0$ & $\textbf{1341.9}\pm\textbf{409.6}$ \\
        \midrule
        Ant - Broken ankle & $1118.7\pm197.7$ & $1094.8\pm208.0$ & $1177.0\pm219.8$ & $\textbf{1364.5}\pm\textbf{311.8}$ \\
        Ant - Broken hip & $2222.2\pm460.2$ & $1998.1\pm421.6$ & $2148.1\pm642.0$ & $\textbf{2534.3}\pm\textbf{275.8}$ \\
        \midrule
        Minitaur - Motor failure & $2.6\pm1.3$ & $3.0\pm1.2$ & $2.7\pm1.7$ & $\textbf{3.1}\pm\textbf{1.2}$ \\
        \bottomrule
    \end{tabular}
    \caption{Adaptation performance under the component damage.}
    \label{tab:damage}
    % \vspace{-1em}
\end{table*}

% \begin{table}[t!]
%     \centering
%     % \footnotesize
%     \captionof{table}{Performance under sensor failures in Ant.}
%     % \vspace{-1em}
%     \label{tab:sensor failure ant}
%         \begin{tabular}{L{2.8em} | R{3.8em} R{3.8em} R{3.8em} R{3.8em}}
%             \toprule  
%             Sensors & Multi & DvD & SMERL & DiR\\
%             \midrule
%             Leg 1 & $1882(258)$ & $1910(196)$ & $1807(249)$ & $\textbf{2210}(\textbf{408})$ \\
            
%             Leg 2 & $1471(233)$ & $\textbf{1675}(\textbf{307})$ & $1398(342)$ & $1295(237)$ \\
            
%             Leg 3 & $1899(323)$ & $1889(183)$ & $1854(269)$ & $\textbf{1976}(\textbf{260})$ \\
            
%             Leg 4 & $1916(350)$ & $2243(250)$ & $2089(436)$ & $\textbf{2502}(\textbf{308})$ \\
%             \bottomrule
%         \end{tabular}
%     % \vspace{-1em}
% \end{table}
\begin{table}[t!]
    \centering
    \footnotesize
    \begin{tabular}{L{2.8em} | R{4.4em} R{4.4em} R{4.4em} R{4.4em}}
        \toprule  
        Sensors & Multi & DvD & SMERL & DiR\\
        \midrule
        Leg 1 & $1882\pm258$ & $1910\pm196$ & $1807\pm249$ & $\textbf{2210}\pm\textbf{408}$ \\
        
        Leg 2 & $1471\pm233$ & $\textbf{1675}\pm\textbf{307}$ & $1398\pm342$ & $1295\pm237$ \\
        
        Leg 3 & $1899\pm323$ & $1889\pm183$ & $1854\pm269$ & $\textbf{1976}\pm\textbf{260}$ \\
        
        Leg 4 & $1916\pm350$ & $2243\pm250$ & $2089\pm436$ & $\textbf{2502}\pm\textbf{308}$ \\
        \bottomrule
    \end{tabular}
    \captionof{table}{Performance under sensor failures in Ant.}
    % \vspace{-1em}
    \label{tab:sensor failure ant}
    
    % \vspace{-1em}
\end{table}

\section{Related Work}
In this work, we focus on the generalization across environments with various dynamics. A common approach to solve this problem is through domain randomization~\cite{tobin2017domain,peng2018sim}, where a single policy is trained under various dynamics in simulation. Prior works have shown the effectiveness of domain randomization for the adaptation across dynamics~\cite{rajeswaran2016epopt,yu2017preparing,akkaya2019solving,shi2022reinforcement,mehta2020active}. Another line of work resolves the generalization through domain adaptation~\cite{chebotar2019closing,hwangbo2019learning,ramos2019bayessim}, which grounds the simulator with the collected transitions from the target domain and train the policy to be optimal under the target dynamics. Furthermore, Robust RL has shown improved transfer performance by optimizing the worst-case performance in the source environment~\cite{pinto2017robust,jiang2021monotonic,mankowitz2019robust}. In contrast, we resolve the generalization using an ensemble of diverse policies.

Searching for diverse solutions has been studied in Evolutionary Computation and Reinforcement Learning research. In Evolutionary Computation, Quality-Diversity~(QD) is a representative type of approach that searches for diverse high-performing solutions~\cite{cully2015robots,mouret2015illuminating,nilsson2021policy}. However, the requirement of defining behavior descriptors limits QD to complicated tasks~\cite{grillotti2022unsupervised}. In Reinforcement Learning, unsupervised skill discovery has been proposed to train a latent conditional policy without environment reward~\cite{eysenbach2018diversity,Sharma2020Dynamics-Aware,hartikainen2019dynamical}, which can prevent the behavior from being practically feasible by ignoring the environment reward. When extrinsic rewards are also considered, several approaches have been proposed to train diverse high-performing policies~\cite{kumar2020one,parker2020effective,masood2019diversity,zhou2022continuously,lupu2021trajectory,zahavy2021discovering,zhang2019learning}. We remark that our open-ended training manner resembles the prior works that train diverse policies iteratively~\cite{zhou2022continuously,zhang2019learning}. However, we focus on controllable diversity through the exhaustive solution discovery different from the methods. Furthermore, several approaches resolve the generalization over various dynamics with the assistance of diverse policies~\cite{kumar2020one,osa2021discovering,kaushik2022safeapt}, same as our work. However, we take the first step to regulate the diversity for more efficient adaptation as far as we know.
 
Our proposed diversity optimization through the inverse dynamics disagreement also resembles Model-based RL~(MBRL)~\cite{deisenroth2019pilco,chua2018deep}. The divergence between inverse dynamics models has also been proposed in the prior work for imitation learning from the observation~\cite{yang2019imitation}. For generalization across dynamics, recent work has achieved the generalization of the dynamics model~\cite{lee2020context,seo2020trajectory}. Unlike these works, we utilize the inverse dynamics models for diverse solution discovery.

\section{Conclusion}
In this work, we present \textit{Diversity in Regulation}~(DiR), a novel diversity-driven algorithm that learns multiple high-performing policies iteratively for adaptation under dynamics variations. The key ingredient of our method is the novel diversity objective through the inverse dynamics disagreement with the state filtration function. Specifically, we can regulate the diversity by customizing the state filtration function for desired behavior patterns. Our empirical results show that DiR can adapt to various test conditions and outperforms prior diversity-driven approaches. Overall, we believe our approach would further strengthen the understanding of diverse solution discovery and could be helpful in safe adaptation under dynamics variations which is critical for the Sim2Real problem.

\section{Acknowledgments}
This research was supported in part by Shanghai Municipal Science and Technology Major Project (No.2021SHZDZX0103), in part by Ji Hua Laboratory, Foshan, China (No.X190011TB190), in part by Science and Technology Development Center, Ministry of Education (No.2021ITA10013).

\bibliography{aaai23}

\clearpage
\section{Appendix}

\section{A. Details of the Motivation Experiment}
\label{apd:motivation}
\begin{figure}[H]
    \centering
    \includegraphics[width=0.2\textwidth]{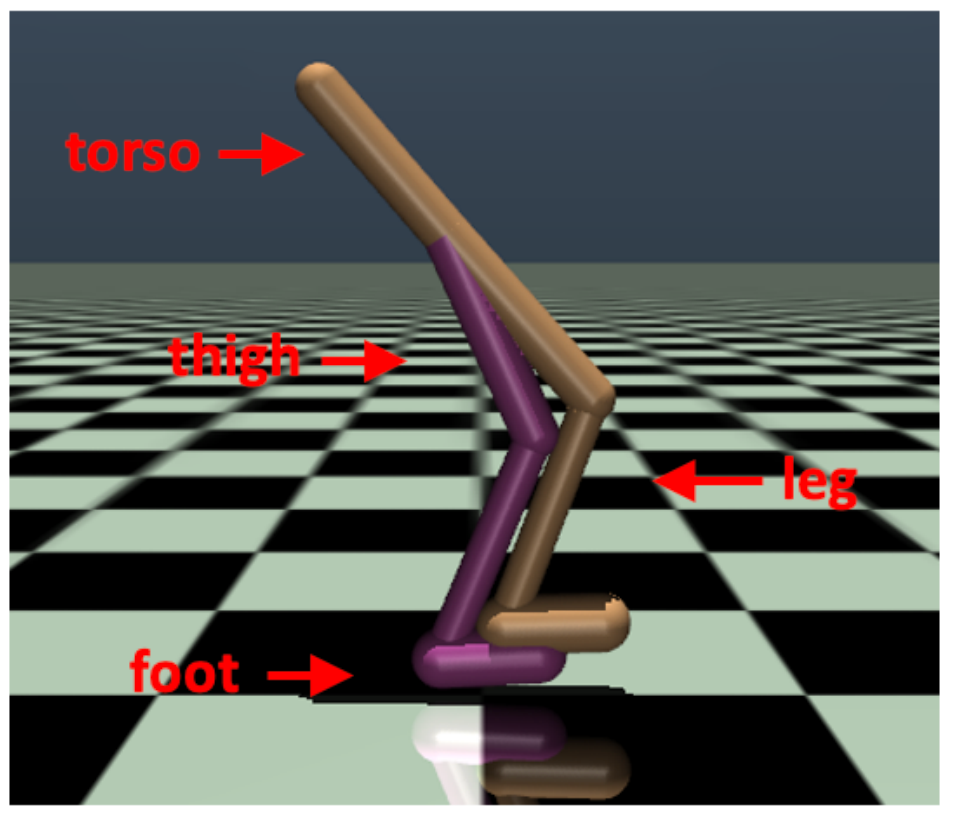}
    \caption{The body components of Walker.}
    \label{fig:walker_body}
\end{figure}

The inverse dynamics models $\{\mathcal{T}_{\phi_i}(a|s,s')\}_{i=1}^4$ have 2 hidden layers with 256 neurons, and output the mean and standard deviation of the action inference given consecutive state pair $(s,s')$. The original state information and the utilized missing state information settings are shown in Table~\ref{tab:missing state settings}. The body components of Walker2D in Mujoco~\cite{todorov2012mujoco} are shown in Figure~\ref{fig:walker_body}.

The training data for inverse dynamics models are collected by a SAC agent. We train the SAC algorithm for $1\mathrm{M}$ time steps. The policy network and twin value functions are 2-hidden-layer MLPs with 128 neurons. We set the learning rate as $0.0003$, batch size as $256$, $\gamma$ as $0.99$, and target update ratio $\tau$ as $0.001$. We train four inverse dynamics models and the SAC agent simultaneously at each step. After training, we collected 30 episodes with the trained policy to compare the inference variance of the IDMs.

\begin{table}[htp]
    \centering
    \caption{The state information settings utilized to train the four inverse dynamics models.}
    \label{tab:missing state settings}
    \begin{tabular}{c|c|c|c}
        \toprule  
        Original state~/ IDM 1 & IDM 2 & IDM 3 & IDM 4   \\  
        \midrule
        pos of torso~($\mathbb{R}^2$) & $\surd$ & $\surd$ & $\surd$  \\  
        \midrule
        pos of 3 left joints~($\mathbb{R}^3$) & $\surd$ & $\surd$ & $\surd$ \\  
        \midrule
        pos of 3 right joints~($\mathbb{R}^3$) & $\surd$ & $\surd$ & $\surd$ \\
        \midrule
        global linear vel~($\mathbb{R}^3$) & $\surd$ & $\surd$ & $\surd$   \\
        \midrule
        ang vel of thigh joints~($\mathbb{R}^2$) & $\surd$ & $\surd$ & $\times$   \\
        \midrule
        ang vel of leg joints~($\mathbb{R}^2$) & $\surd$ & $\times$ & $\times$   \\
        \midrule
        ang vel of foot joints~($\mathbb{R}^2$) & $\times$ & $\times$ & $\times$   \\
        \bottomrule
    \end{tabular}
\end{table}

\section{B. Extended Discussion of DiR}
\subsection{B.1~Derivation of the Diversity Objective with Inverse Dynamics Models}
\begin{manualtheorem}{B.1}
    \label{threm:diversity with idm}
    Assume the inference divergence of the trained inverse dynamics models are bounded and minor:
    \begin{equation*}
        \left|\mathbb{E}\left[\log\dfrac{\mathcal{T}_{\phi_i}(a|f(s),f(s'))}{\mathrm{P}^{\pi_i}(a|f(s),f(s'))}\right]\right|\leq \epsilon,~\forall~1\leq i\leq M, 
    \end{equation*}
    the diversity objective:
    \begin{align*}
        &Div(\pi_k,\pi_i)=\\
        &\mathbb{E}_{\pi_k}\left[\mathrm{D}_{KL}\left[\mathrm{P}^{\pi_k}(\cdot|f(s),f(s'))\parallel\mathrm{P}^{\pi_i}(\cdot|f(s),f(s'))\right]\right]
    \end{align*}
    from Eq.~\ref{eq:lagrangian} can be approximately lower bounded by the objective
    \begin{align*}
        Div_{ce}(\pi_k,\pi_i)=\mathbb{E}_{(s,a,s')\sim \pi_k}\left[-\log \mathcal{T}_{\phi_i}(a|f(s),f(s'))\right],
    \end{align*}
    from Eq.~\ref{eq:new div} with less entropy reduction of $\pi_k$.\\
    
    \noindent \textbf{Proof}.
    We denote the $f(s)$ as $\bar{s}$ for simplicity.
    \begin{align*}
        Div&(\pi_k,\pi_i)=\mathbb{E}_{(s,s')\sim \pi_k}\left[\mathrm{D}_{KL}\left[
        \mathrm{P}^{\pi_k}(\cdot|\bar{s},\bar{s}')\parallel\mathrm{P}^{\pi_i}(\cdot|\bar{s},\bar{s}')\right]\right]\\
        &=\mathbb{E}_{(s,s')\sim \pi_k}\left[\int_a \mathrm{P}^{\pi_k}(a|\bar{s},\bar{s}')\log\mathrm{P}^{\pi_k}(a|\bar{s},\bar{s}')\right]\\
        &\quad~ +\mathbb{E}_{(s,s')\sim \pi_k}\left[-\int_a\mathrm{P}^{\pi_k}(a|\bar{s},\bar{s}')\log\mathrm{P}^{\pi_i}(a|\bar{s},\bar{s}')\right]\\
        &=\mathbb{E}_{(s,s')\sim \pi_k}\left[D_{KL}\left[\mathrm{P}^{\pi_k}\parallel\mathcal{T}_{\phi_k}\right]\right] \\
        &\quad~+\mathbb{E}_{(s,s')\sim\pi_k}\left[\mathbb{E}_{a\sim\mathrm{P}^{\pi_k}}\left[\log\mathcal{T}_{\phi_k}(a|\bar{s},\bar{s}')\right]\right]\\
        &\quad~-\mathbb{E}_{(s,s')\sim \pi_k}\left[\mathbb{E}_{a\sim \mathrm{P}^{\pi_k}}\left[\log\mathrm{P}^{\pi_i}(a|\bar{s},\bar{s}')\right]\right]\\
        &\geq \mathbb{E}_{(s,s')\sim \pi_k}\left[\mathbb{E}_{a\sim \mathrm{P}^{\pi_k}}\left[\log\mathcal{T}_{\phi_k}(a|\bar{s},\bar{s}')\right]\right]\\
        &\quad -\mathbb{E}_{(s,s')\sim \pi_k}\left[\mathbb{E}_{a\sim \mathrm{P}^{\pi_k}}\left[\log\mathrm{P}^{\pi_i}(a|\bar{s},\bar{s}')\right]\right]\quad~(KL\geq0)\\
    \end{align*}
    since maximizing the first term of RHS above will result in entropy reduction of policy $\pi_k$ which is harmful for exploration, we propose to omit the first term and further derive the RHS as:
    \begin{align*}
        RHS &= -\mathbb{E}_{(s,s')\sim\pi_k}\left[\mathbb{E}_{a\sim\mathrm{P}^{\pi_k}}\left[\log\mathrm{P}^{\pi_i}(a|\bar{s},\bar{s}')\right]\right]\\
        &= -\mathbb{E}_{(s,a,s')\sim\pi_k}[\log\mathrm{P}^{\pi_i}(a|\bar{s},\bar{s}')]\qquad\qquad\quad~(Eq.\ref{eq:id})\\
        &= -\mathbb{E}_{(s,a,s')\sim\pi_k}[\log\mathrm{P}^{\pi_i}(a|\bar{s},\bar{s}') - \log\mathcal{T}_{\phi_i}(a|\bar{s},\bar{s}') \\
        &\quad~ + \log\mathcal{T}_{\phi_i}(a|\bar{s},\bar{s}')]\\
        &=-\mathbb{E}_{(s,a,s')\sim\pi_k}\left[\log\mathcal{T}_{\phi_i}(a|\bar{s},\bar{s}')\right] \\
        &\quad~ +\mathbb{E}_{(s,a,s')\sim\pi_k}\left[\log\dfrac{\mathrm{P}^{\pi_i}(a|\bar{s},\bar{s}')}{\mathcal{T}_{\phi_i}(a|\bar{s},\bar{s}')}\right]\\
        &\approx -\mathbb{E}_{(s,a,s')\sim\pi_k}\left[\log\mathcal{T}_{\phi_i}(a|\bar{s},\bar{s}')\right]\qquad(Assumption)\\
        &=\mathbb{E}_{(s,a,s')\sim \pi_k}\left[-\log \mathcal{T}_{\phi_i}(a|f(s),f(s'))\right]\\
        &=Div_{ce}(\pi_k,\pi_i)
    \end{align*}
\end{manualtheorem}

\hfill

\subsection{B.2 Algorithms}

\begin{algorithm}[H]

    \caption{Diversity in Regulation}
    
    \label{alg:dir}
    {\bf Input:} Population size $M$, number of training iterations $N$, state filtration function $f(s)$ , initial policies $\{\pi_{\theta_i}(a|s)\}_{i=1}^M$, initial inverse dynamics models $\left\{\mathcal{T}_{\phi_i}(a|f(s),f(s'))\right\}_{i=1}^M$, scaling factor $\alpha$.\\
    {\bf Output:} Diverse policies $\{\pi_{\theta_i}(a|s)\}_{i=1}^M$.
    \begin{algorithmic}[1]
        \For{k = 1, 2, \dots, M}
            \For{j = 1, 2, \dots, N}
                \State Collect trajectories with $\pi_k$.
                
                \If{k $\geq$ 2}
                    \State Refine the rewards from the trajectories as:
                    $$
                    r_t = r_t + \dfrac{\alpha}{k - 1}\sum_{i=1}^{k-1}\left[-\log \mathcal{T}_{\phi_i}(a_t|f(s_t),f(s_{t+1}))\right]
                    $$
                \EndIf
                
                \State Train $\pi_k$ with collected trajectories using PPO.
                
                \State Train $\mathcal{T}_{\phi_k}$ with collected trajectories by maximizing the log-likelihood.
            \EndFor
        \EndFor
    \end{algorithmic}
\end{algorithm}

\begin{algorithm}[H]
    \caption{Few-shot adaptation}
    \label{alg:adap}
    {\bf Input:} Policies $\{\pi_i\}_{i=1}^M$, test environment $\mathcal{M}$, population size $M$. number of test episodes $N$.\\
    {\bf Output:} The best-performing policy $\pi^*$.
    \begin{algorithmic}[1]
        \For{k = 1, 2, \dots, M}
            \State Rollout policy $\pi_{\theta_k}$ in $\mathcal{M}$ for $N$ episodes and calculate averaged episodic rewards $S_k$.
            \State Get the index of the best performing policy $ b = \arg\max_{j\in\{1,\dots,M\}}S_j$.
            \State $\pi^*\leftarrow \pi_b$.
        \EndFor
    \end{algorithmic}
\end{algorithm}

\begin{table*}[!tb]
    \centering
    \caption{The original state information and the removed information through the state filtration functions for all four environments. $\surd$ suggests the function does not remove the information, and $\times$ indicates removing the information.}
    \label{tab:state filtration}
    \begin{tabular}{cccc}
        \toprule  
        Ant & Hopper & Walker & Minitaur   \\  
        \midrule
        pos of torso~($\mathbb{R}^1$) - $\surd$ 
        & pos of torso~($\mathbb{R}^2$) - $\surd$ 
        & pos of torso~($\mathbb{R}^2$) - $\surd$ 
        & pos of FL leg~($\mathbb{R}^2$) - $\times$   \\  
        \midrule
        pos of leg 1~($\mathbb{R}^3$) - $\times$ 
        & pos of 3 joints~($\mathbb{R}^3$) - $\times$ 
        & pos of 3 left joints~($\mathbb{R}^3$) - $\surd$ 
        & pos of BL leg~($\mathbb{R}^2$) - $\times$  \\  
        \midrule
        pos of leg 2~($\mathbb{R}^3$) - $\times$ 
        & global linear vel~($\mathbb{R}^3$) - $\surd$ 
        & pos of 3 right joints~($\mathbb{R}^3$) - $\times$ 
        & pos of FR leg~($\mathbb{R}^2$) - $\times$ \\
        \midrule
        pos of leg 3~($\mathbb{R}^3$) - $\times$ 
        & ang vel of 3 joints~($\mathbb{R}^3$) - $\times$ 
        & global linear vel~($\mathbb{R}^3$) - $\surd$ 
        & pos of BR leg~($\mathbb{R}^2$) - $\times$    \\
        \midrule
        pos of leg 4~($\mathbb{R}^3$) - $\times$ 
        & N/A 
        & ang vel of 3 left joints~($\mathbb{R}^3$) - $\surd$ 
        & ang vel of FL leg~($\mathbb{R}^2$) - $\surd$  \\
        \midrule
        linear vel of torso~($\mathbb{R}^3$) - $\surd$ 
        & N/A 
        & ang vel of 3 right joints~($\mathbb{R}^3$) - $\times$ 
        & ang vel of BL leg~($\mathbb{R}^2$) - $\surd$  \\
        \midrule
        ang vel of torso~($\mathbb{R}^3$) - $\surd$  
        & N/A 
        & N/A 
        & ang vel of FR leg~($\mathbb{R}^2$) - $\surd$   \\
        \midrule
        ang vel of leg 1~($\mathbb{R}^2$) - $\times$  
        & N/A 
        & N/A 
        & ang vel of BR leg~($\mathbb{R}^2$) - $\surd$   \\
        \midrule
        ang vel of leg 2~($\mathbb{R}^2$) - $\times$
        & N/A 
        & N/A 
        & torques of FL leg~($\mathbb{R}^2$) - $\surd$   \\
        \midrule
        ang vel of leg 3~($\mathbb{R}^2$) - $\times$
        & N/A 
        & N/A 
        & torques of BL leg~($\mathbb{R}^2$) - $\surd$    \\
        \midrule
        ang vel of leg 4~($\mathbb{R}^2$) - $\times$ 
        & N/A 
        & N/A 
        & torques of FR leg~($\mathbb{R}^2$) - $\surd$   \\
        \midrule
        contact forces~($\mathbb{R}^{84}$) - $\surd$ 
        & N/A 
        & N/A 
        & torques of BR leg~($\mathbb{R}^2$) - $\surd$   \\
        \midrule
        N/A
        & N/A 
        & N/A 
        & pos of torso~($\mathbb{R}^4$) - $\surd$   \\
        \bottomrule
    \end{tabular}
\end{table*}

\section{C. Environments and Implementation Details}

\subsection{C.1 Test Environments with Dynamics Mismatch}

\begin{figure}[h]
    \centering
    \includegraphics[width=0.45\textwidth]{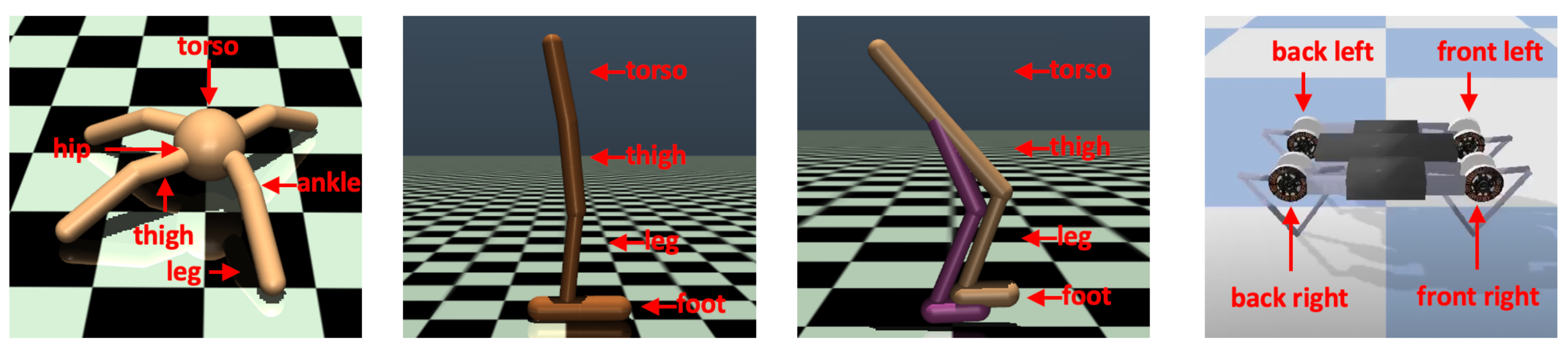}
     \caption{All environments and the detailed body components.}
    \label{fig:components}
\end{figure}

To validate the adaptation performance of the approaches, we implement various test circumstances with dynamics shifts. The detailed body components of all environments are shown in Figure~\ref{fig:components}. We implement a set of environments with different damaged components as follows:
\begin{itemize}
    \item \textbf{Ant - broken ankle}: disabled ankle joint on the first leg.
    \item \textbf{Ant - broken hip}: disabled hip joint on the first leg.
    \item \textbf{Hopper - broken leg}: disabled joint connecting the thigh and the leg.
    \item \textbf{Hopper - broken foot}: disabled joint connecting the leg and the foot.
    \item \textbf{Walker - broken leg}: disabled joint connecting the thigh and the leg on the right side of the torso.
    \item \textbf{Walker - broken foot}: disabled joint connecting the leg and the foot on the right side of the torso.
    \item \textbf{Minitaur - motor failure}: disabled motors on the left back leg.
\end{itemize}

In addition, we also evaluate the approaches under the perturbation of dynamics parameters~(e.g.~mass). Here we implement test environments as follows:
\begin{itemize}
    \item \textbf{Ant - leg mass}: the masses of four legs are scaled according to different magnitudes.
    \item \textbf{Ant - ankle friction}: the frictions of four ankle joints are scaled according to different magnitudes.
    \item \textbf{Hopper - foot mass}: the mass of the foot is scaled according to different magnitudes.
    \item \textbf{Hopper - foot friction}: the friction of the foot joint is scaled according to different magnitudes.
    \item \textbf{Walker - foot mass}: the masses of two feet are scaled according to different magnitudes.
    \item \textbf{Walker - foot friction}: the frictions of two feet joints are scaled according to different magnitudes.
\end{itemize}

Finally, we implement conditions with sensor failures to evaluate the robustness of the policies in the preference of missing observations as follows:
\begin{itemize}
    \item \textbf{Ant - leg $k$ sensor}: the sensors detecting the positions of joints on leg k are in failure, and the corresponding state values are always zeros.
    \item \textbf{Walker - left/right leg sensor}: the sensors detecting the positions of joints on the left/right leg are in failure, and the corresponding state values are always zeros.
\end{itemize}

\subsection{C.2 Details of State Filtration Functions}~
To assist the policy population in discovering diverse locomotion patterns, we focus on the diversity of the leg motions. Thus, we design the state filtration functions to remove partial information about the legs. The detailed state information and the filtration functions are shown in Table~\ref{tab:state filtration}.

\subsection{C.3 Implementation Details}

We implement all algorithms with PPO~\cite{schulman2017proximal} backbone. We implement the policy networks as 2-hidden MLP with 64 neurons and the value network as 2-hidden MLP with 128 neurons. We utilize the tanh function as the activation. We set the learning rate as $3e^{-4}$, batch size as $256$, discount rate $\gamma$ as $0.99$, GAE parameter $\lambda$ as $0.95$, entropy coefficient as $0.1$, value loss coefficient as $1$, clipping parameter as $0.25$, and number of epochs as $10$.

\noindent\textbf{DiR}~~For the inverse dynamics models, we utilize 2-hidden MLP with 128 neurons. We set the tradeoff coefficient $\alpha$ as $0.05$ for Hopper and Walker, $0.01$ for Ant, $0.0005$ for Minitaur.

\noindent\textbf{SMERL}~~We implement the discriminator as 2-hidden MLP with 256 neurons. SMERL only optimizes the diversity objective when the trajectory reward exceeds the threshold. We set the trajectory threshold ratio $\epsilon=0.1$ and intrinsic reward coefficient $\alpha=10$, the same as the original paper.

\noindent\textbf{DvD}~~Since the determinant diversity objective in the original paper is computationally inefficient for training 10 policies simultaneously, we modify the determinant diversity objective to an intrinsic reward 
$$r_{in}=\dfrac{1}{M-1}\sum_{j\neq k}^M\left[\exp\left(-\dfrac{\parallel \pi_k(a_t|s_t) - \pi_j(a_t|s_t) \parallel^2}{2}\right)\right]$$ 
which is the same as the diversity measurement in the original paper. For the tradeoff coefficient, we perform a grid search over $0.5$, $0.1$, $0.05$, $0.01$, $0.005$, $0.001$. The coefficient is fixed to $0.1$ for Hopper and Walker, $0.01$ for Ant, and $0.001$ for Minitaur.

\noindent\textbf{Multi}~~We train 10 policies with different initial parameters simultaneously, and we train the policies without the diversity optimization.

\noindent\textbf{PG}~~We train one single policy with the Vanilla PPO.

\section{D. Additional Results}

\subsection{D.1 Selected Policies in Test Environments}

Here we analyze the performance of policies in the test environments to examine whether different policies are selected given various test conditions. We present the explicit performance of all policies under the broken component circumstances, as shown in Tables~\ref{tab:adap_walker},~\ref{tab:adap_hopper},~\ref{tab:adap_ant}. The results demonstrate that different policies are selected under different test conditions, validating that DiR learns multiple policies with different locomotion patterns. When unpredictable circumstances happen in the environment, DiR can select the best-performing policy from the ensemble for efficient adaptation.

\subsection{D.2 Performance under Sensor Failures}
As we train regulated diverse policies by introducing the state filtration functions, we hypothesize that the policies output actions relying on different state variables from the removed information. Thus, we implement test environments with various sensor failures where the state variables are always zeros and report the performance of the diverse policies in the environments. Here we implement four test environments where the sensors receiving leg positions are in failure for Ant and two test environments for Walker. We report the maximum performance of the policies in the test environments over 8 random seeds. As shown in Tables~\ref{tab:sensor failure ant},~\ref{tab:sensor failure walker}, DiR achieves better or competitive performance compared with baseline methods under different sensor failure conditions, which validates that the policies output actions conditioning on different state variables of the state information.
\begin{table}[H]
    \centering
    \footnotesize
    \captionof{table}{Adaptation performance under sensor failures in Walker.}
        \label{tab:sensor failure walker}
        \vspace{-1em}
        \begin{tabular}{L{3.2em}|R{4.4em} R{4.4em} R{4.4em} R{4.4em}}
            \toprule  
            Sensors & Multi & DvD & SMERL & DiR\\
            \midrule
            Left L & $2566\pm469$ & $2346\pm423$ & $2152\pm547$ & $\textbf{2646}\pm\textbf{586}$ \\
            
            Right L & $\textbf{2842}\pm\textbf{316}$ & $2612\pm237$ & $2502\pm225$ & $\textbf{2718}\pm\textbf{356}$ \\
            \bottomrule
        \end{tabular}
\end{table}

\begin{table*}[h!]
    \centering
    \caption{Adaptation performance and the selected policies in test environments of Walker.}\label{tab:adap_walker}
    \begin{tabular}{c|cccccccccc}
        \toprule\toprule
        Test Env & policy 1 & policy 2 & policy 3 & policy 4 & policy 5 & policy 6 & policy 7 & policy 8 & policy 9 & policy 10 \\
       \midrule
        broken foot & \textbf{2258.69} & 1003.17 & 117.91 & 27.51 & -3.19 & 1039.94 & -3.00 & 293.91 & 993.35 & 1037.74 \\
        broken leg  & 472.49 & \textbf{2822.81} & 2690.13 & 984.79 & 509.47 & 961.89 & 801.23 & 740.60 & 2911.41 & 2588.05 \\ 
        \bottomrule\bottomrule
    \end{tabular}
\end{table*}

\begin{table*}[h!]
    \centering
    \caption{Adaptation performance and the selected policies in test environments of Hopper.}\label{tab:adap_hopper}
    \begin{tabular}{c|cccccccccc}
        \toprule\toprule
        Test Env & policy 1 & policy 2 & policy 3 & policy 4 & policy 5 & policy 6 & policy 7 & policy 8 & policy 9 & policy 10 \\
       \midrule
        broken foot & 998.98 & 558.02 & \textbf{999.15} & 993.77 & 10.23 & 998.24 & 998.26 & 27.21 & 998.41 & 995.99  \\
        broken leg  & 712.34 & 460.93 & 1027.29& \textbf{3130.23} & 201.54 & 2108.46 & 241.32 & 191.59 & 268.51 & 339.16 \\ 
        \bottomrule\bottomrule
    \end{tabular}
\end{table*}
    
\begin{table*}[h!]
    \centering
    \caption{Adaptation performance and the selected policies in test environments of Ant.}\label{tab:adap_ant}
    \begin{tabular}{c|cccccccccc}
        \toprule\toprule
        Test Env & policy 1 & policy 2 & policy 3 & policy 4 & policy 5 & policy 6 & policy 7 & policy 8 & policy 9 & policy 10 \\
       \midrule
        broken ankle & 617.12 & 596.33 & 248.10 & 1116.29 & 694.70 & 176.31 & \textbf{821.59} & 474.9 & 342.97 & 666.11 \\
        broken hip  & 625.28 & 1507.77 & 732.94 & 678.78 & 1280.11 & \textbf{2265.75} & 1465.87 & 2179.59 & 1899.25 & 1489.05 \\ 
        \bottomrule\bottomrule
    \end{tabular}
\end{table*}

\subsection{D.3 Visualization of Discovered Behaviors}

\begin{figure*}[h!]
    \vspace{-1em}
    \centering
    \begin{subfigure}{\textwidth}
        \centering
        \includegraphics[width=0.7\textwidth]{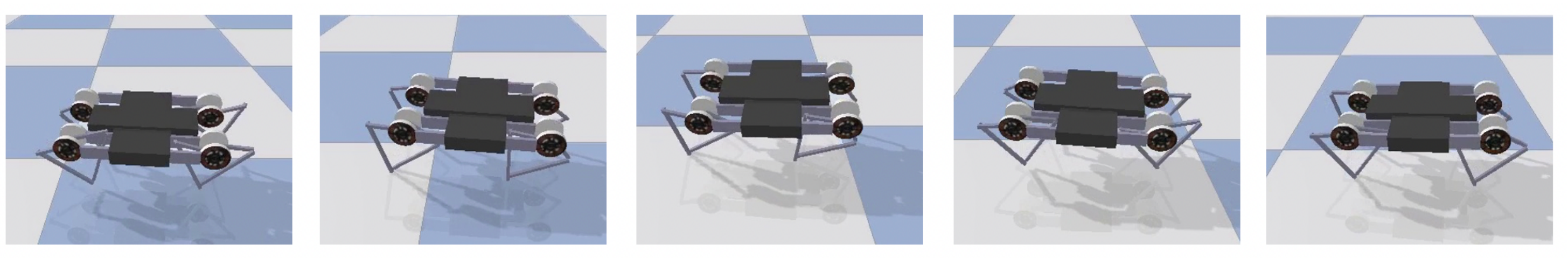}
        \caption{Running steady.}
    \end{subfigure}
    
    \begin{subfigure}{\textwidth}
        \centering
        \includegraphics[width=0.7\textwidth]{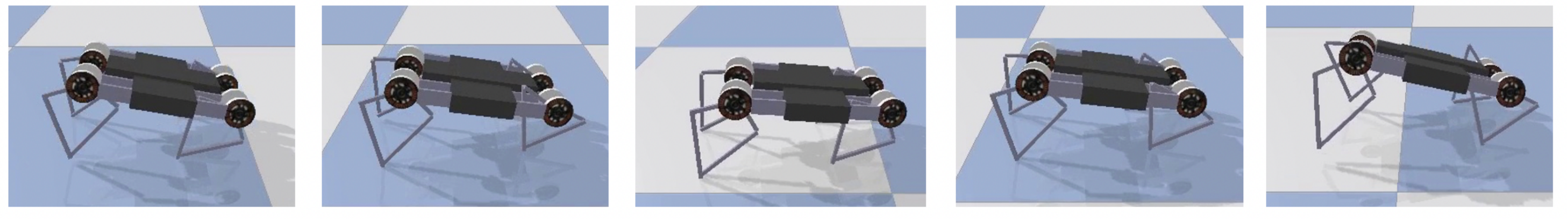}
        \caption{Running with head down.}
    \end{subfigure}
    
    \begin{subfigure}{\textwidth}
        \centering
        \includegraphics[width=0.7\textwidth]{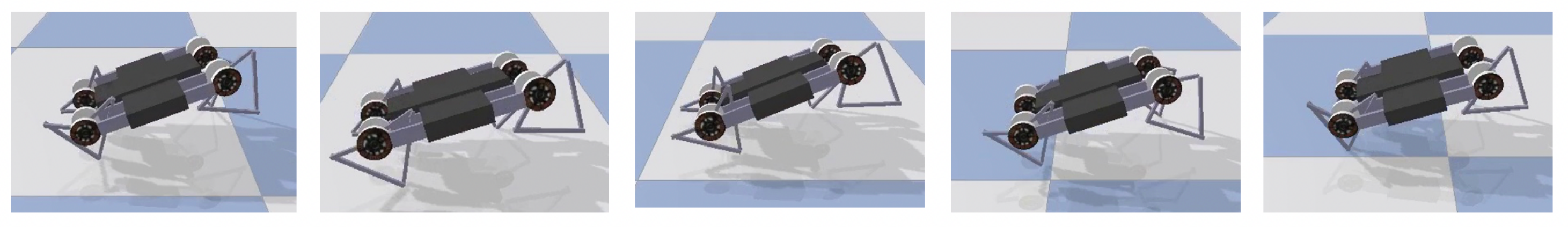}
        \caption{Running with head up.}
    \end{subfigure}
    
    \caption{Behaviors in Minitaur.}
    \vspace{-1em}
\end{figure*} 

\begin{figure*}
    \vspace{-1em}
    \centering
    \begin{subfigure}{\textwidth}
        \centering
        \includegraphics[width=0.65\textwidth]{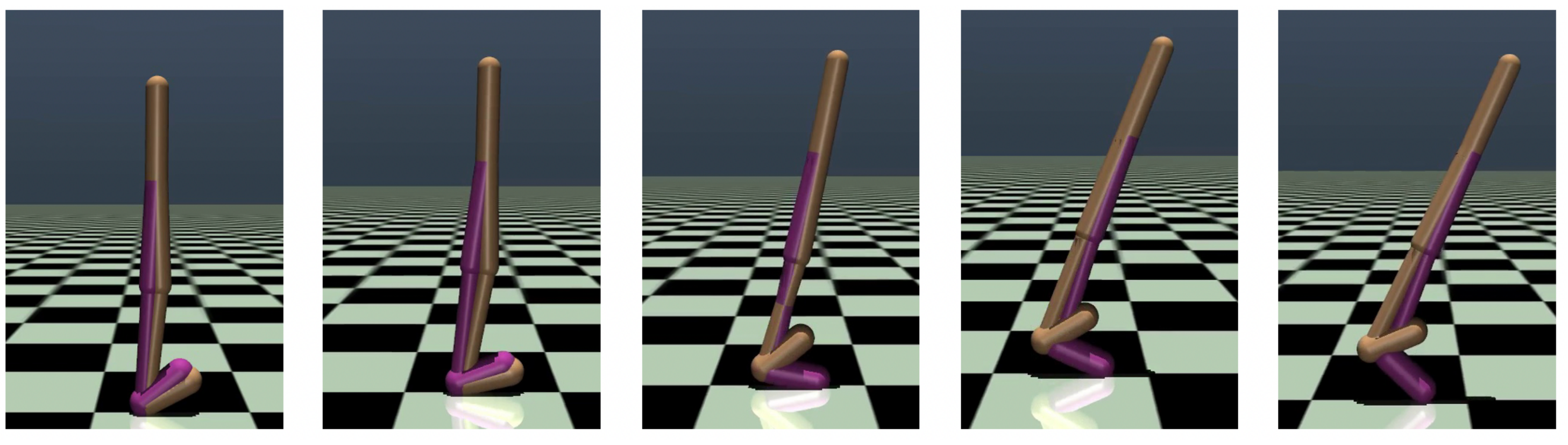}
        \caption{Jumping with the right foot.}
    \end{subfigure}
    
    \begin{subfigure}{\textwidth}
        \centering
        \includegraphics[width=0.65\textwidth]{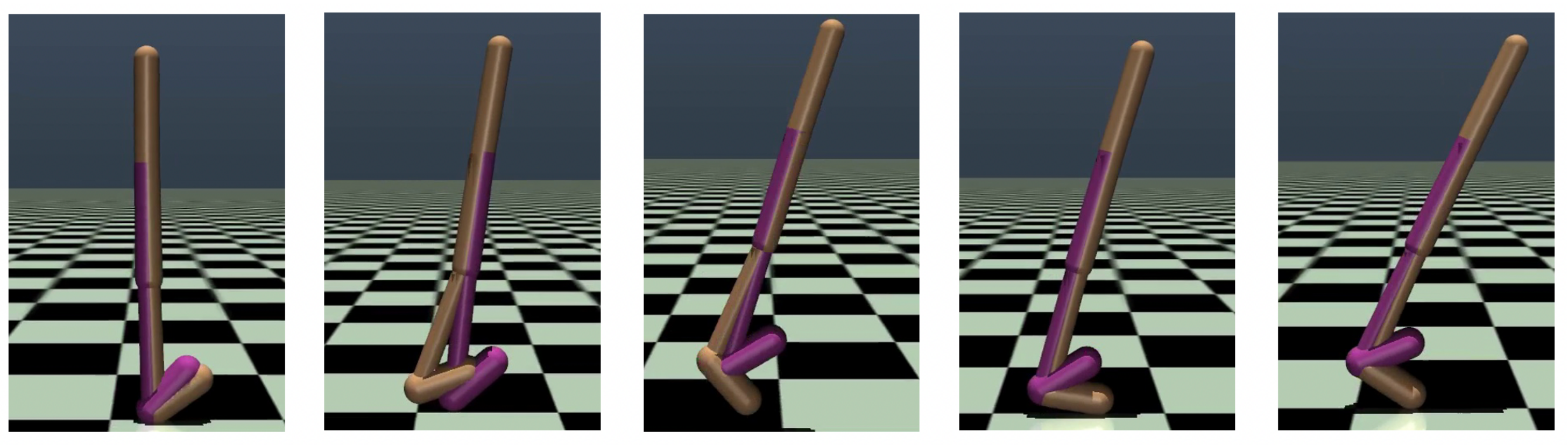}
        \caption{Jumping with the left foot.}
    \end{subfigure}
    
    \begin{subfigure}{\textwidth}
        \centering
        \includegraphics[width=0.65\textwidth]{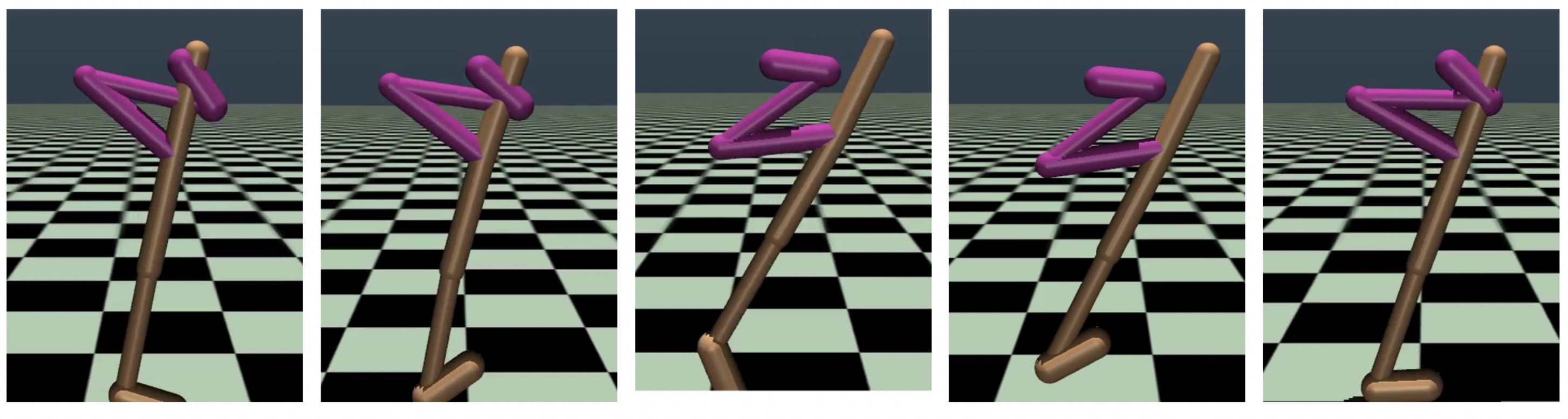}
        \caption{Jumping with one foot raising up feet.}
    \end{subfigure}
    
    \caption{Behaviors in Walker.}
    \vspace{-1em}
\end{figure*} 

\begin{figure*}
    \vspace{-1em}
    \centering
    \begin{subfigure}{\textwidth}
        \centering
        \includegraphics[width=0.65\textwidth]{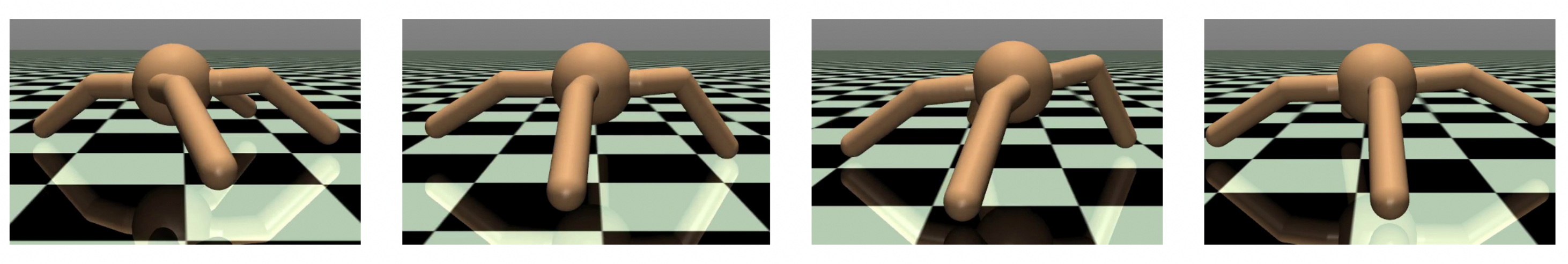}
        \caption{Running with two legs.}
    \end{subfigure}
    
    \begin{subfigure}{\textwidth}
        \centering
        \includegraphics[width=0.65\textwidth]{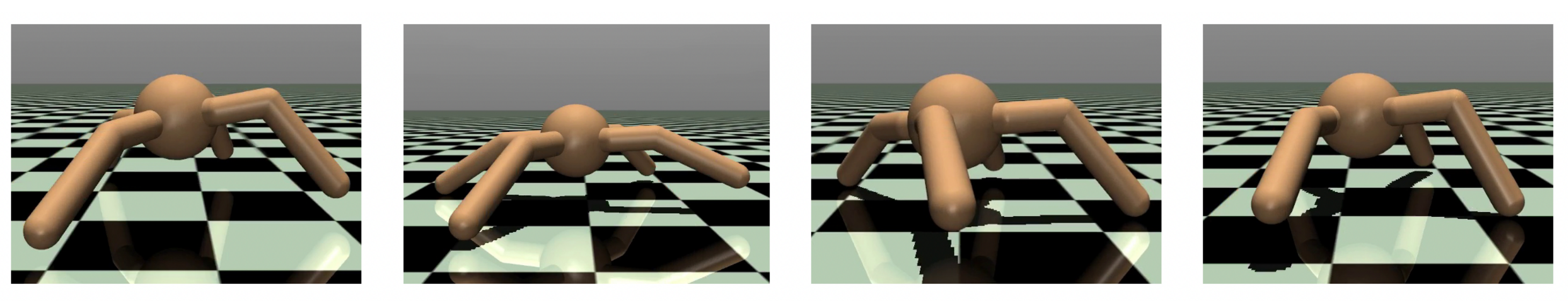}
        \caption{Walking with four legs.}
    \end{subfigure}
    
    \caption{Behaviors in Ant.}
    \vspace{-1em}
\end{figure*}

\begin{figure*}
    \vspace{-1em}
    \centering
    \begin{subfigure}{\textwidth}
        \centering
        \includegraphics[width=0.65\textwidth]{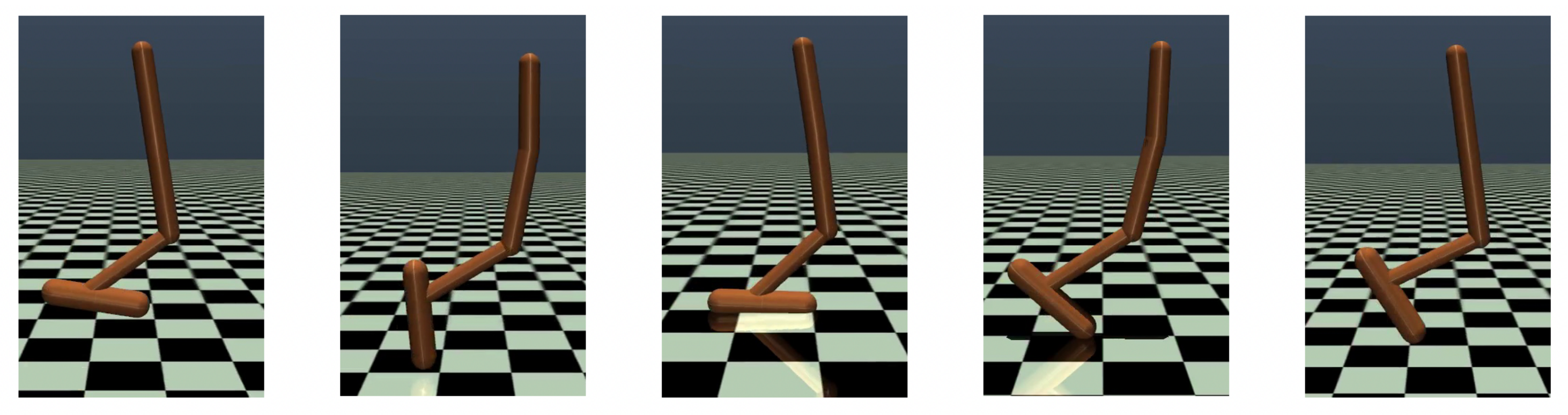}
        \caption{Hopping with torso bent.}
    \end{subfigure}
    
    \begin{subfigure}{\textwidth}
        \centering
        \includegraphics[width=0.65\textwidth]{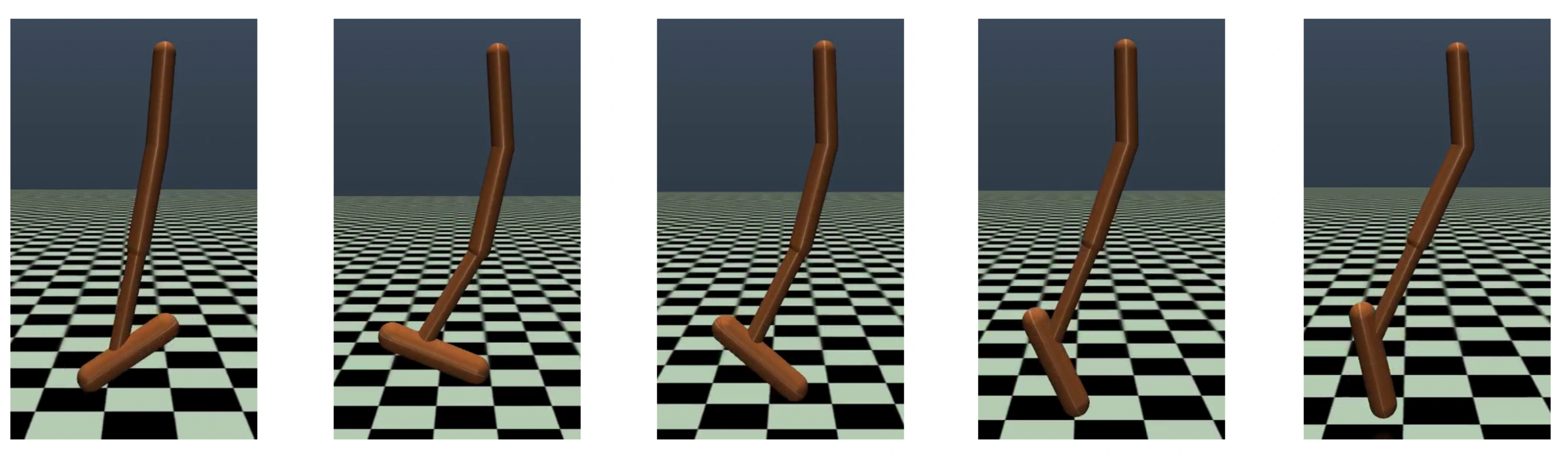}
        \caption{Hopping high.}
    \end{subfigure}
    
    \begin{subfigure}{\textwidth}
        \centering
        \includegraphics[width=0.65\textwidth]{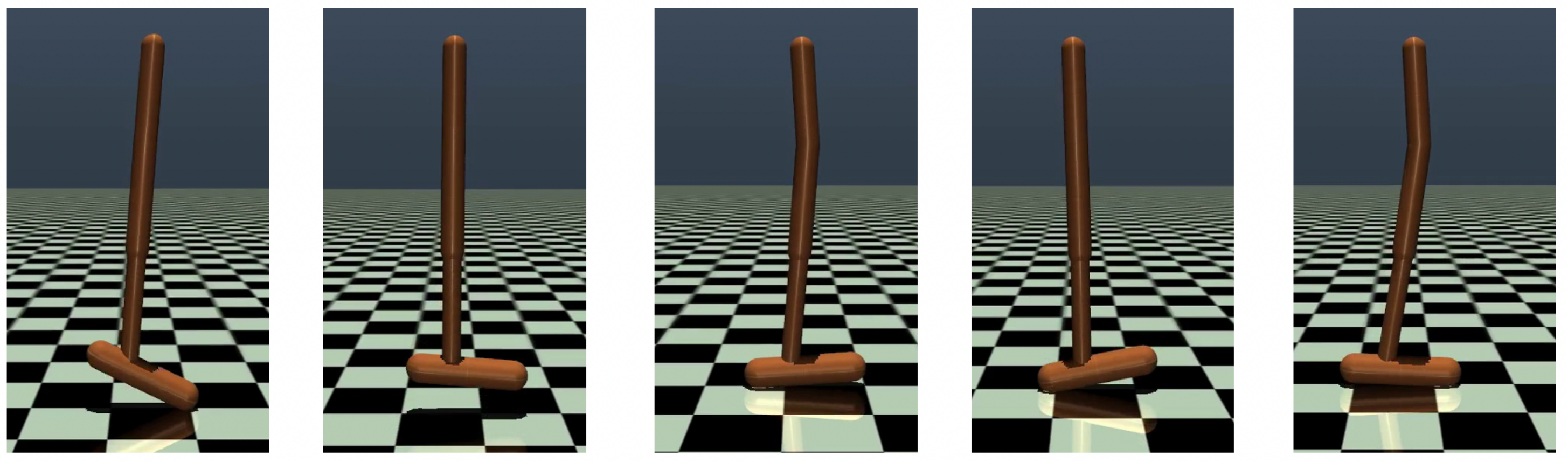}
        \caption{Hopping vertically in place.}
    \end{subfigure}
    
    \caption{Behaviors in Hopper}
    \vspace{-1em}
\end{figure*}

\clearpage

\end{document}